\pdfoutput=1

\documentclass[11pt]{article}

\usepackage[preprint]{acl}

\usepackage{times}
\usepackage{latexsym}

\usepackage[T1]{fontenc}

\usepackage[utf8]{inputenc}

\usepackage{microtype}

\usepackage{inconsolata}

\usepackage{graphicx}
\usepackage{booktabs}
\usepackage{multirow}
\usepackage{colortbl}
\usepackage[table]{xcolor}
\usepackage{arydshln}
\usepackage{makecell}
\usepackage{adjustbox}
\usepackage{array}
\usepackage{easyReview}
\usepackage{pifont}
\usepackage{booktabs}
\usepackage{pifont}
\usepackage{array}
\usepackage{amsmath}
\usepackage{enumitem}
\usepackage{xcolor}
\usepackage{tabularx}
\newcommand{\cmark}{\color{green}\ding{51}}%
\newcommand{\xmark}{\color{red}\ding{55}}%
\usepackage{wasysym}
\usepackage[normalem]{ulem}

\definecolor{cadblue}{HTML}{1024FF}

\definecolor{lightgrayrow}{RGB}{245,245,245}
\definecolor{lightbluerow}{RGB}{230,245,255}
\definecolor{lightbluehighlight}{RGB}{242,249,253}

\definecolor{PrussianBlue}{RGB}{30,144,255}

\title{\textsc{ExpConCAD}: Experience-Guided Text-to-CAD Generation from Shape Descriptions with Implicit Spatial Constraints}

\author{%
\begin{tabular}{c}
Jingyao Liu\textsuperscript{1,2} \quad
Jinkang Tang\textsuperscript{1,2} \quad
Chen Huang\textsuperscript{1,2,3} \quad
Wenqiang Lei\textsuperscript{1,2}\footnotemark[1] \quad
See-Kiong Ng\textsuperscript{3}
\end{tabular}
\vspace{0.5em}
\\
\begin{tabular}{c}
\small \textsuperscript{1}College of Computer Science, Sichuan University \\
\small \textsuperscript{2}Engineering Research Center of Machine Learning and Industry Intelligence, \\
\small Ministry of Education, China \\
\small \textsuperscript{3}Institute of Data Science, National University of Singapore \\
\small \texttt{liujingyao1@stu.scu.edu.cn}
\end{tabular}
}

\begin{document}
\maketitle
\begin{abstract}
Text-to-CAD aims to generate executable CAD programs from natural-language descriptions. However, real-world descriptions are often underspecified and omit critical spatial constraints required for valid CAD construction, a challenge that has been largely overlooked by existing methods. In this paper, we argue that missing spatial constraints should be inferred with respect to the underlying construction structure and informed by reusable design experience. Based on this insight, we propose \textsc{ExpConCAD}, an experience-enhanced framework for implicit spatial constraint completion. \textsc{ExpConCAD} first recovers the intended construction structure and constraint scopes, then retrieves relevant constraint-completion experience for similar scopes to complete the missing spatial constraints, and finally generates executable CadQuery programs. Extensive experiments demonstrate the effectiveness of \textsc{ExpConCAD} and provide insights into the role of construction structure understanding and experience memory in spatial constraint completion. Our code is available at: \url{https://github.com/Hotjiashell/ExpConCAD}.
\end{abstract}

\section{Introduction}
\label{sec:intro}

\begin{figure*}[t]
    \centering
    \includegraphics[width=\textwidth]{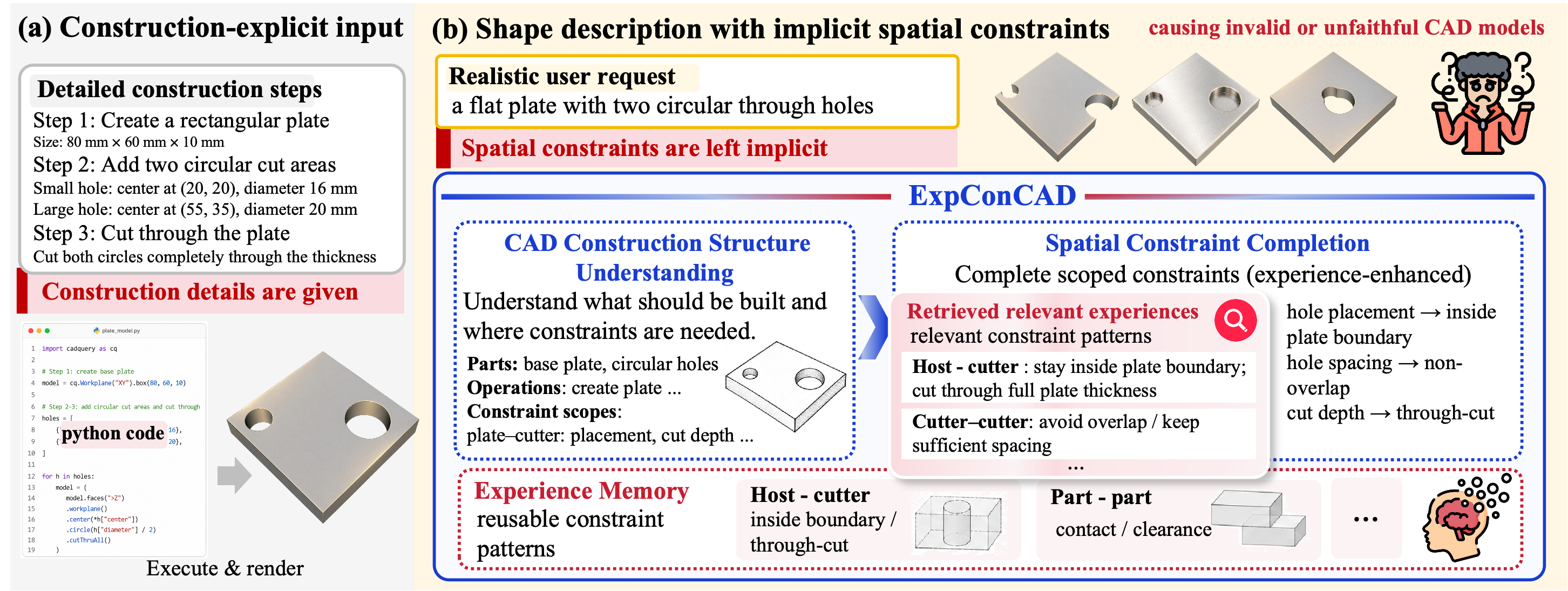}
\caption{Overview of \textsc{ExpConCAD} for resolving implicit spatial constraints (Fig.a and Fig.b).
\textsc{ExpConCAD} first recovers the construction structure to identify constraint scopes, and then completes missing spatial constraints with guidance from transferable experience.}
\vspace{-3mm}
    \label{fig:intro}
\end{figure*}

Automatically translating natural-language design descriptions into executable CAD construction programs, such as CadQuery scripts\cite{cadquery}, allows users to create 3D models directly from textual instructions without manually operating professional CAD software~\citep{khan2024text2cad,CADTranslator}. Recent advances in large language models (LLMs)~\citep{openai2025gpt5,qwen2025qwen3,alibaba2026qwen35} have substantially improved this Text-to-CAD paradigm\citep{badagabettu2024query2cad,wang2026text2cad}. As shown in Figure \ref{fig:intro}(a), given a detailed specification of a target object, the LLM generates the Python code, which is then executed to construct the corresponding 3D geometry.

However, existing Text-to-CAD methods typically require users to explicitly specify detailed geometric parameters, modeling operations, and spatial relations of the target object~\citep{khan2024text2cad,CADTranslator,liao2025automated,2025cadrille,guan2025cadcoder,govindarajan2025cadmium}. While such fine-grained inputs help define precise spatial constraints among object components, specify such over-detailed instructions impose substantial burden on users. In realistic scenarios, users usually describe the desired shape at a much higher semantic level. For example, as shown in Figure~\ref{fig:intro}(b), a user may request ``\textit{a flat plate with two circular through holes}''. Such descriptions convey the intended appearance of the object, while leaving much of the underlying spatial constraints implicit. As a result, existing methods often generate structurally incorrect CAD models, such as overlapping holes. Recent study attempts to address this under-specification through end-to-end optimization by directly aligning under-specified inputs with target CAD shapes \citep{wang2025cadfusion}. However, it only learns input–output correlations while overlooking the underlying implicit spatial constraints. Consequently, when encountering unseen designs with different spatial arrangements or component interactions, it still struggles to generalize and fails to infer the missing constraints for accurate CAD generation. Therefore, \textbf{effectively reasoning about \uline{implicit spatial constraints} in under-specified instructions remains a critical challenge for generating accurate 3D CAD structures}.

To address this challenge, we start from a basic property of CAD construction: building a valid CAD model first requires identifying geometric elements and modeling operations, and then specifying the spatial relations among them~\citep{kyratzi2021constraint,camba2016parametric}. 
Thus, for underspecified shape descriptions with implicit constraints, the missing spatial relations should be completed with respect to the underlying construction structure and its constraint scopes, rather than inferred from surface descriptions alone.
We further observe that similar constraint-completion problems repeatedly appear across historical CAD construction processes~\citep{kyratzi2021constraint,camba2016parametric}. 
For example, a cylindrical part with side holes and a plate with through holes have different target geometries, but both involve comparable constraint scopes, such as cutter--base placement and cutter--cutter separation. 
This suggests that prior CAD cases can provide transferable experience, such as cutter containment, cutter separation, and through-cut depth, for completing missing constraints over similar scopes.
Inspired by these observations, we propose \textbf{\textsc{ExpConCAD}} (\textbf{\uline{Exp}}erience-enhanced \textbf{\uline{Con}}straint completion for \textbf{\uline{CAD}}), a framework for generating faithful CAD programs from underspecified descriptions by explicitly completing implicit spatial constraints. 
\textsc{ExpConCAD} couples \textbf{construction-aware spatial constraint reasoning} with an \textbf{experience memory}. 
The reasoning pipeline first performs \textit{Construction Structure Understanding} to recover the intended construction process and identify constraint scopes, then performs \textit{Spatial Constraint Completion} before generating executable CadQuery code. 
The experience memory distills reusable constraint-completion patterns from prior CAD cases and retrieves relevant experience to guide completion over similar scopes.

We evaluate \textsc{ExpConCAD} on \textsc{CadFusion-Hard}, a challenging benchmark for Text-to-CAD with underspecified spatial constraints. 
\textsc{ExpConCAD} improves VLM-Score over the strongest baseline by 22.5\% on Qwen3.5-27B and 27.9\% on GPT-5. 
Further analyses show that Construction Structure Understanding provides reliable scopes for Spatial Constraint Completion, while experience memory is most effective when it stores reusable scoped constraint-completion patterns.
Our contributions are three-fold:
\begin{itemize}[leftmargin=*,itemindent=0.05cm, itemsep=-2pt]
    \item We highlight the \textit{implicit spatial constraint} as a critical challenge in Text-to-CAD, where natural descriptions often omit spatial relations required for valid CAD construction.
    \item We propose \textsc{ExpConCAD}, which combines construction-aware spatial constraint reasoning and reusable experience memory from prior CAD cases for improved CadQuery generation.

    \item We conduct extensive experiments to validate our effectiveness and characteristics. 


\end{itemize}
\section{Related Work}
\label{sec:related_work}

\begin{table*}[t]
\centering
\footnotesize
\setlength{\tabcolsep}{3.2pt}
\renewcommand{\arraystretch}{1.10}

\begin{tabularx}{\textwidth}{@{}l c c c >{\raggedright\arraybackslash}X@{}}
\toprule
\textbf{Method}
& \textbf{ISC}
& \textbf{Input}
& \textbf{Output}
& \textbf{Core Idea} \\
\midrule

Text2CAD~\citep{khan2024text2cad}
& \xmark
& CED
& Seq.
& Construction-explicit sequence generation. \\

CADRille~\citep{2025cadrille}
& \xmark
& CED
& CQ
& Construction-explicit CadQuery generation. \\

CAD-Coder~\citep{guan2025cadcoder}
& \xmark
& CED
& CQ
& CoT-based operation planning for CadQuery generation. \\

CADFusion~\citep{wang2025cadfusion}
& \xmark
& USD
& Seq.
& Visual supervision for shape alignment. \\

\midrule

\textbf{\textsc{ExpConCAD}}
& \cmark
& \textbf{USD}
& \textbf{CQ}
& \textbf{Experience-guided spatial constraint completion.} \\

\bottomrule
\end{tabularx}

\caption{\textbf{Comparison of representative Text-to-CAD methods.}
ISC indicates whether the method explicitly addresses implicit spatial constraint completion.
CED: construction-explicit descriptions;
USD: underspecified shape descriptions with implicit spatial constraints;
Seq.: CAD command sequence;
CQ: CadQuery code.}
\label{tab:cad_comparison}
\end{table*}

CAD modeling is essential for mechanical design and digital manufacturing, but creating CAD models typically requires domain expertise and familiarity with professional modeling software~\citep{cherng1998feature,shah1995parametric,bhavnani2001beyond,chester2007teaching}. 
Text-to-CAD aims to lower this barrier by converting natural-language design descriptions into structured CAD representations. 
Many existing methods represent CAD models as parametric command sequences~\citep{2021DeepCAD,khan2024text2cad,CADTranslator,liao2025automated,wang2025cadgpt}, which often require dedicated interpreters or reconstruction procedures to recover CAD models. 
Recent work has therefore shown growing interest in generating executable CAD programs, such as CadQuery scripts~\citep{cadquery,xie2025texttocadquery,2025cadrille,guan2025cadcoder}, which is executed to create CAD models. 
Following this direction, we focus on {generating executable CadQuery programs from natural-language descriptions}.

However, most Text-to-CAD methods assume that the input description explicitly specifies the information required for CAD construction, including modeling operations, geometric parameters, and spatial relations. Under this assumption, both task-specific models~\citep{khan2024text2cad,CADTranslator,liao2025automated,xie2025texttocadquery,2025cadrille,guan2025cadcoder} and recent LLM-based approaches~\citep{cadcodeverify,yuan2026procad,schupbach2025text,li2025llm4cad} primarily focus on translating detailed user instructions into target CAD representations. However, real-world descriptions are often underspecified and omit critical spatial constraints required for CAD construction. As a result, generating executable CAD programs from such descriptions requires not only translation, but also inferring the missing constraints. Only limited work has explored this problem. CADFusion~\citep{wang2025cadfusion} leverages visual supervision to improve shape consistency, but it mainly aligns generated shapes with textual descriptions rather than explicitly reasoning about the missing spatial constraints. Consequently, it struggles to infer implicit constraints and generalize to underspecified inputs. Therefore, completing missing spatial constraints from incomplete natural-language descriptions remains a largely unexplored challenge.
\section{\textsc{ExpConCAD}}
\label{sec:method}

\begin{figure*}[t]
    \centering
    \includegraphics[width=\textwidth]{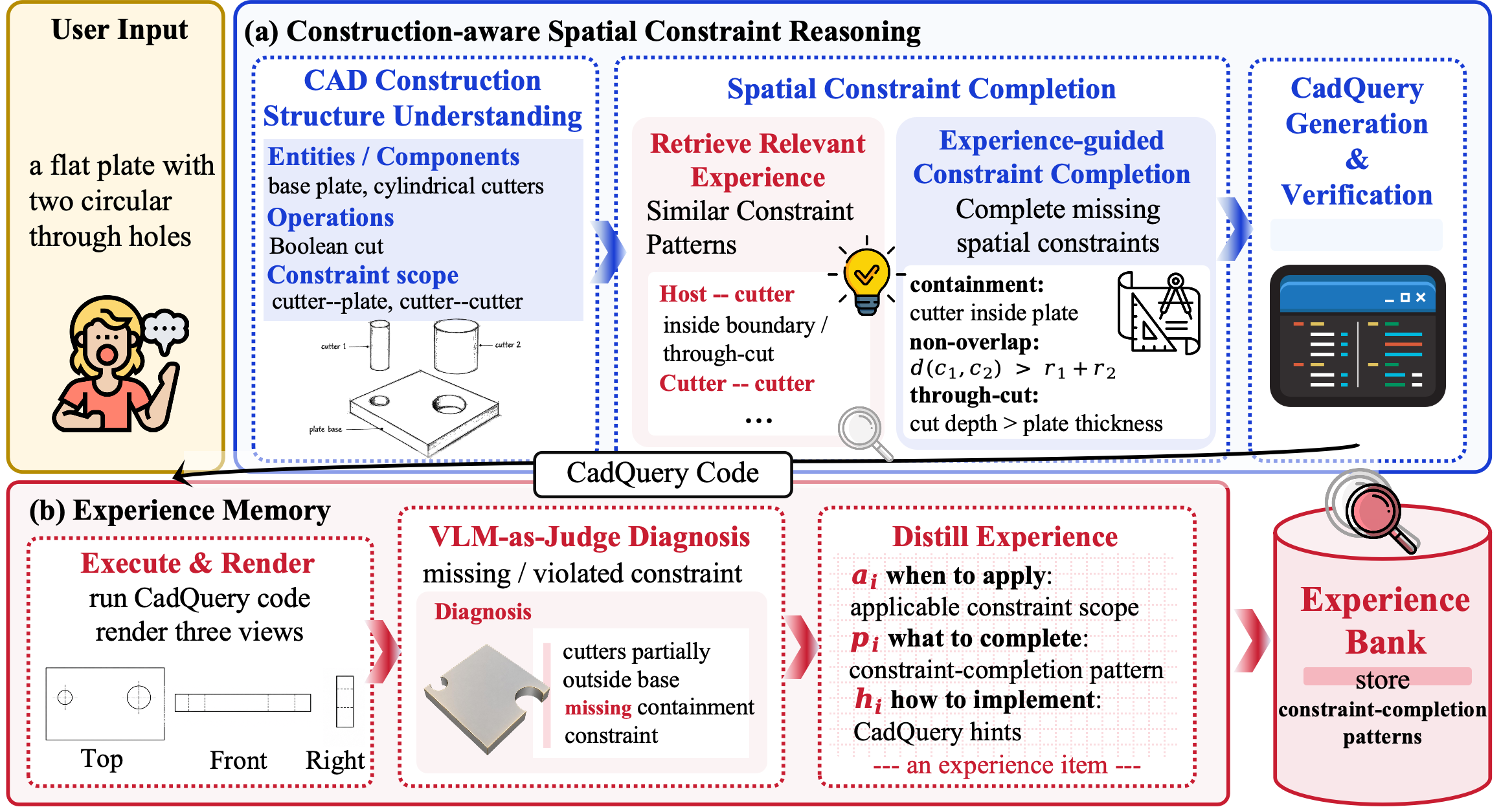}
    \caption{
\textbf{Overview of \textsc{ExpConCAD}.}
\textbf{(a)} The framework identifies constraint scopes through construction-structure recovery and completes implicit spatial constraints before CadQuery generation.
\textbf{(b)} Experience memory stores reusable constraint-completion patterns from prior CAD cases for future retrieval.
}
    \label{fig:method}
\end{figure*}

\noindent\textbf{Problem Formulation and Overview}. Given an underspecified inputs $x$, our goal is to generate an executable CadQuery program $y$. The central challenge is to complete the implicit spatial constraints that are omitted from $x$ but are required for valid CAD construction. 
As shown in Figure~\ref{fig:method}, \textsc{ExpConCAD} addresses this challenge through two key designs: 
(a) a {construction-aware spatial constraint reasoning} pipeline, and 
(b) an {experience memory} $\mathcal{B}$ that contains reusable guidance for constraint completion. 
The reasoning pipeline first recovers the intended construction process and identifys the scopes where spatial constraints should be completed. 
It then performs spatial constraint completion with the support of retrieved experience $\mathcal{R}$ from $\mathcal{B}$. 
Finally, the completed constraints are used to generate the executable CadQuery program.

\subsection{Construction-Aware Spatial Constraint Reasoning}
\label{sec:online_pipeline}

To turn the under-specified descriptions into valid CadQuery programs, as shown in Figure~\ref{fig:method}(a), \textsc{ExpConCAD} involves the following three steps:

\paragraph{Construction Structure Understanding (CSU).}
This step recovers the construction process implied by the input and identifies the scopes for subsequent spatial constraint completion:
\begin{equation}
    s = L_{\mathrm{CSU}}(x).
\end{equation}
The recovered structure $s$ includes the CAD elements, modeling operations, and constraint scopes where missing spatial relations should be completed.
A constraint scope specifies where a missing constraint should hold, such as between an operation and its target object or between two geometric elements.
For example, for ``\textit{a plate with two circular through holes}'', CSU identifies the base plate, two cylindrical cutters, and two Boolean-cut operations, and derives scopes such as cutter--plate placement and cutter--cutter separation from these elements, operations, and the design intent.

\paragraph{Spatial Constraint Completion (SCC).}
This step infers the missing spatial relations within the identified scopes to support valid CAD generation:
\begin{equation}
    c = L_{\mathrm{SCC}}(x,s).
    \label{eq:scc_base}
\end{equation}
The completed constraint set $c$ specifies how the identified elements and operations should be spatially configured, including coordinate conditions, distance bounds, size relations, and operation parameters.
For example, for ``\textit{a plate with two circular through holes}'', SCC instantiates the missing constraints as parameter-level conditions: each cutter center should lie within the plate boundary, the distance between cutter centers should exceed the sum of their radii, and the cut depth should be larger than the plate thickness.
Section~\ref{sec:experience_memory} describes how retrieved experience facilitates this process.

\paragraph{CadQuery Generation.}
After completing the spatial constraints, \textsc{ExpConCAD} generates the final executable program:
\begin{equation}
    y = L_{\mathrm{Code}}(x,s,c).
\end{equation}
The code generator conditions on the original description $x$, the recovered construction structure $s$, and the completed constraint set $c$ to produce executable CadQuery code.
To ensure that the program can produce a CAD model, we execute it for compilation and rendering checks.
If execution fails, only the code is revised according to the error message, while $s$ and $c$ remain fixed to preserve the completed spatial reasoning.

\subsection{Experience Memory for Spatial Constraint Completion}
\label{sec:experience_memory}

We construct a constraint-completion experience memory from prior CAD cases to enhance SCC. For instance, repeated side-hole constructions reveal reusable constraint patterns, including cutter containment, hole non-overlap, and sufficient cutting depth. \textsc{ExpConCAD} distills these patterns into reusable experience and retrieves relevant memories to guide constraint completion in analogous construction scopes.

\paragraph{Experience Format.}
To guide constraint completion over similar constraint scopes, each experience item should be both retrievable and actionable.
Therefore, we design each item to answer three questions: \uline{when} it should be applied, \uline{what} constraint-completion pattern it provides, and \uline{how} the pattern can be implemented in CadQuery.
Formally, each item is represented as
\begin{equation}
    e_i = (a_i, p_i, h_i),
\end{equation}
where $a_i$ specifies the applicable constraint scope used for experience retrieval, including the involved CAD elements and operations;
$p_i$ describes the reusable constraint-completion pattern, such as containment, non-overlap, symmetry, or through-cut;
and $h_i$ records how to implement these constraints in CadQuery, such as parameter bounds, placement rules, coordinate conditions, or Boolean-operation settings.
For example, a cutter--plate experience item may use $a_i$ to specify a hole-cutting scope between a cutter and a base solid, $p_i$ to describe boundary-containment and through-cut constraints, and $h_i$ to record CadQuery implementation hints such as valid coordinate bounds and cut-depth settings.

\paragraph{Experience Construction.}
We construct experience items from failed prior generations, which reveal missing or violated spatial constraints that the model has not yet handled.
For each generated CadQuery program $y_i$, we execute the code and render the resulting CAD model into views $v_i$.
Given the description $x_i$ and rendered views $v_i$, a VLM-as-judge diagnoses constraint-related errors:
\begin{equation}
    d_i = L_{\mathrm{Judge}}(x_i, v_i).
\end{equation}
The diagnosis focuses on missing or violated spatial constraints, such as overlapping holes, cutters outside the base, or cuts that do not pass through the object.
We distill experience only from diagnosed failure cases, using the diagnosis and the corresponding implementation context:
\begin{equation}
    e_i = L_{\mathrm{Distill}}(d_i, y_i) = (a_i, p_i, h_i).
\end{equation}
For example, if a plate with two through holes contains overlapping holes, the failure can be distilled into a cutter--cutter experience item for non-overlap, with center-distance bounds as implementation hints.
If a cutter is placed outside the plate or fails to cut through it, the failure can be distilled into a cutter--plate item for boundary containment or through-cut depth.
This turns instance-level failures into reusable experience to guide constraint completion in analogous construction scopes.

\paragraph{Experience Use.}
Experience memory enhances the basic SCC process in Eq.~\ref{eq:scc_base} with relevant experience from similar constraint scopes.
During inference, given $x$ and the recovered construction structure $s$, we construct fine-grained queries over the identified constraint scopes:
\begin{equation}
    Q=\{q_j\}_{j=1}^{m}=L_{\mathrm{Query}}(x,s).
\end{equation}
Each query corresponds to a scope where constraints may be missing, such as cutter--plate placement or cutter--cutter separation in the plate example.
We retrieve relevant experience by matching each query with the applicable construction situation $a_i$ of each memory item:
\begin{equation}
    \mathcal{R}=\bigcup_{j=1}^{m}\mathrm{TopK}(q_j,\mathcal{B}).
\end{equation}
This retrieves experience from similar constraint scopes rather than whole CAD cases.
Spatial Constraint Completion is then enhanced by conditioning on both the current construction structure and retrieved experience:
\begin{equation}
    c = L_{\mathrm{SCC}}(x,s,\mathcal{R}).
\end{equation}
Thus, the memory provides targeted guidance for completing constraints within the current scopes.

\section{Experiments}
\label{sec:experiments}

\begin{table*}[t]
    \centering
    \footnotesize
    \setlength{\tabcolsep}{2.8pt}
    \renewcommand{\arraystretch}{1.10}
    
    \begin{adjustbox}{width=0.95\textwidth}
    \begin{tabular}{lcccccccc}
        \toprule
        \multirow{2}{*}{\textbf{Method}} & \multicolumn{8}{c}{\textbf{CadFusion-Hard}} \\
        \cmidrule(lr){2-9}
        & \textbf{VLM$\uparrow$}
        & \textbf{m. r-IoU$\uparrow$}
        & \textbf{md. r-IoU$\uparrow$}
        & \textbf{m. CD$\downarrow$}
        & \textbf{md. CD$\downarrow$}
        & \textbf{m. rCD$\downarrow$}
        & \textbf{md. rCD$\downarrow$}
        & \textbf{Inv.$\downarrow$} \\
        \midrule
        \rowcolor{lightgrayrow} \multicolumn{9}{c}{\textit{Trained Text-to-CAD Models}} \\
        Text2CAD & 1.88 & 0.182 & 0.140 & 104.983 & 90.268 & 70.634 & 51.074 & \textbf{0.00} \\
        Cadrille & 1.17 & 0.042 & 0.021 & 156.687 & 142.039 & 83.456 & 74.495 & 5.00 \\
        CAD-Coder & 1.54 & 0.283 & 0.248 & 67.414 & 51.511 & 39.109 & 27.750 & 3.00 \\
        CadFusion & 4.22 & 0.287 & 0.230 & 65.582 & 67.797 & 31.310 & 26.268 & 23.50 \\
        \midrule
        \rowcolor{lightgrayrow} \multicolumn{9}{c}{\textit{Qwen3.5-27B}} \\
        Vanilla & 3.33 & 0.233 & 0.112 & 88.289 & 123.459 & 39.586 & 54.98 & 50.50 \\
        CoT & 3.05 & 0.206 & 0.133 & 102.622 & 127.411 & 42.674 & 52.875 & 67.00 \\
        CADCodeVerify & 5.11 & 0.324 & 0.300 & 59.125 & 42.948 & 26.607 & 18.654 & 13.50 \\
        \arrayrulecolor{black!35}\hdashline\arrayrulecolor{black}
        \rowcolor{lightbluehighlight} \textbf{Ours w/o Exp.} & \underline{5.63} & \underline{0.342} & \textbf{0.333 }& \textbf{50.264} & \textbf{32.729} & \underline{22.484} & \textbf{14.435} & \underline{1.00} \\
        \rowcolor{lightbluerow} \textbf{Ours} & \textbf{6.26} & \textbf{0.343} & \underline{0.313} & \underline{51.952} & \underline{34.638} & \textbf{21.925} & \underline{14.525} & 2.00 \\
        \midrule
        \rowcolor{lightgrayrow} \multicolumn{9}{c}{\textit{GPT-5}} \\
        Vanilla & 4.09 & 0.276 & 0.195 & 84.995 & 103.535 & 34.609 & 33.499 & 38.50 \\
        CoT & 4.05 & 0.284 & 0.229 & 81.875 & 69.961 & 28.787 & 25.695 & 32.00 \\
        CADCodeVerify & 5.16 & 0.326 & 0.281 & 62.673 & 47.859 & 26.449 & 20.651 & 21.00 \\
        \arrayrulecolor{black!35}\hdashline\arrayrulecolor{black}
        \rowcolor{lightbluehighlight} \textbf{Ours w/o Exp.} & \underline{6.14} & \underline{0.361} & \underline{0.348} & \underline{50.178} & \underline{32.568} & \underline{20.514} & \underline{13.348} & \underline{0.50} \\
        \rowcolor{lightbluerow} \textbf{Ours} & \textbf{6.60} & \textbf{0.371} & \textbf{0.351} & \textbf{47.496} & \textbf{29.487} & \textbf{19.431} & \textbf{11.122} & \underline{0.50} \\
        \bottomrule
    \end{tabular}
    \end{adjustbox}

\caption{Main results on \textsc{CadFusion-Hard}. 
Best and second-best results are shown in \textbf{bold} and \underline{underlined}. 
Non-renderable outputs are included with worst-decile scores estimated from valid outputs.}
    \label{tab:cadfusion-hard-main}
\end{table*}

\subsection{Experimental Setup}
\label{sec:exp_setup}

\noindent\textbf{Benchmarks and Experience Construction.}
We report main results on \textsc{CADFusion-Hard}, the more challenging subset of CADFusion~\citep{wang2025cadfusion} with underspecified shape descriptions.
Full CADFusion results and \textsc{CadFusion-Hard} construction details are provided in Appendices~\ref{app:full-cadfusion-results} and~\ref{app:benchmark}.
Following prior experience-based settings~\citep{cai2025trainingfreegrpo,ouyang2026reasoningbank}, we construct the experience bank from randomly sampled training examples, keeping all evaluation samples disjoint. 
We report the strongest experience-enhanced result for each backbone and provide a detailed analysis of how experience-bank size affects performance in Section~\ref{sec:analysis_exp_size}.
Details are in Appendix~\ref{app:construct_exp}.

\noindent\textbf{Baselines and Backbones.}
We compare with baselines covering both input-description settings discussed in Section~\ref{sec:related_work}.
The baselines include trained Text-to-CAD models, i.e., \uline{Text2CAD}~\citep{khan2024text2cad}, \uline{CADRille}~\citep{2025cadrille}, \uline{CAD-Coder}~\citep{guan2025cadcoder}, and \uline{CADFusion}~\citep{wang2025cadfusion}, as well as LLM-based methods, including \uline{Vanilla}, \uline{CoT}, and the agentic baseline \uline{CadCodeVerify}~\citep{cadcodeverify}.
We also report \uline{Ours w/o Exp.} to isolate the effect of experience.
For LLM-based methods, since some baselines require both textual and rendered-view inputs, we use two \uline{multimodal LLMs} from different model families, \uline{Qwen3.5-27B} and \uline{GPT-5}, as backbones.
Implementation details are provided in Appendix~\ref{app:baseline_details}.

\noindent\textbf{Metrics.}
Following prior Text-to-CAD evaluations~\citep{khan2024text2cad,wang2025cadfusion}, we report \uline{VLM-Score (VLM)}, \uline{rotation-invariant Intersection-over-Union (r-IoU)}, \uline{Chamfer Distance (CD)}, \uline{rotation-invariant Chamfer Distance (rCD)}, and \uline{Invalid Ratio (Inv.)}.
VLM measures text-shape consistency; r-IoU, CD, and rCD measure geometric quality via volumetric overlap and point-cloud distance; Inv. measures generation validity.
For r-IoU, CD, and rCD, we report both mean (m.) and median (md.) values.
Since these metrics are typically computed only on successfully rendered outputs, large Inv. differences in the main comparison may unfairly favor methods with many invalid CAD models.
We therefore include non-renderable outputs in the main aggregation using a worst-decile value estimated from valid outputs, motivated by validity-aware evaluation~\citep{mallis2026text}.
For analysis experiments, where Inv. differences are less pronounced, we use standard rendered-only computation.
(Appendices~\ref{app:main_res}, \ref{app:diagnostic_settings}, and~\ref{app:metrics})

\noindent\textbf{Implementation Details}. All $L_{\cdot}$ modules are implemented with LLM prompting. Detailed prompts are provided in Appendix~\ref{app:expconcad_details}.

\begin{figure*}[t]
    \centering
    \includegraphics[width=\textwidth]{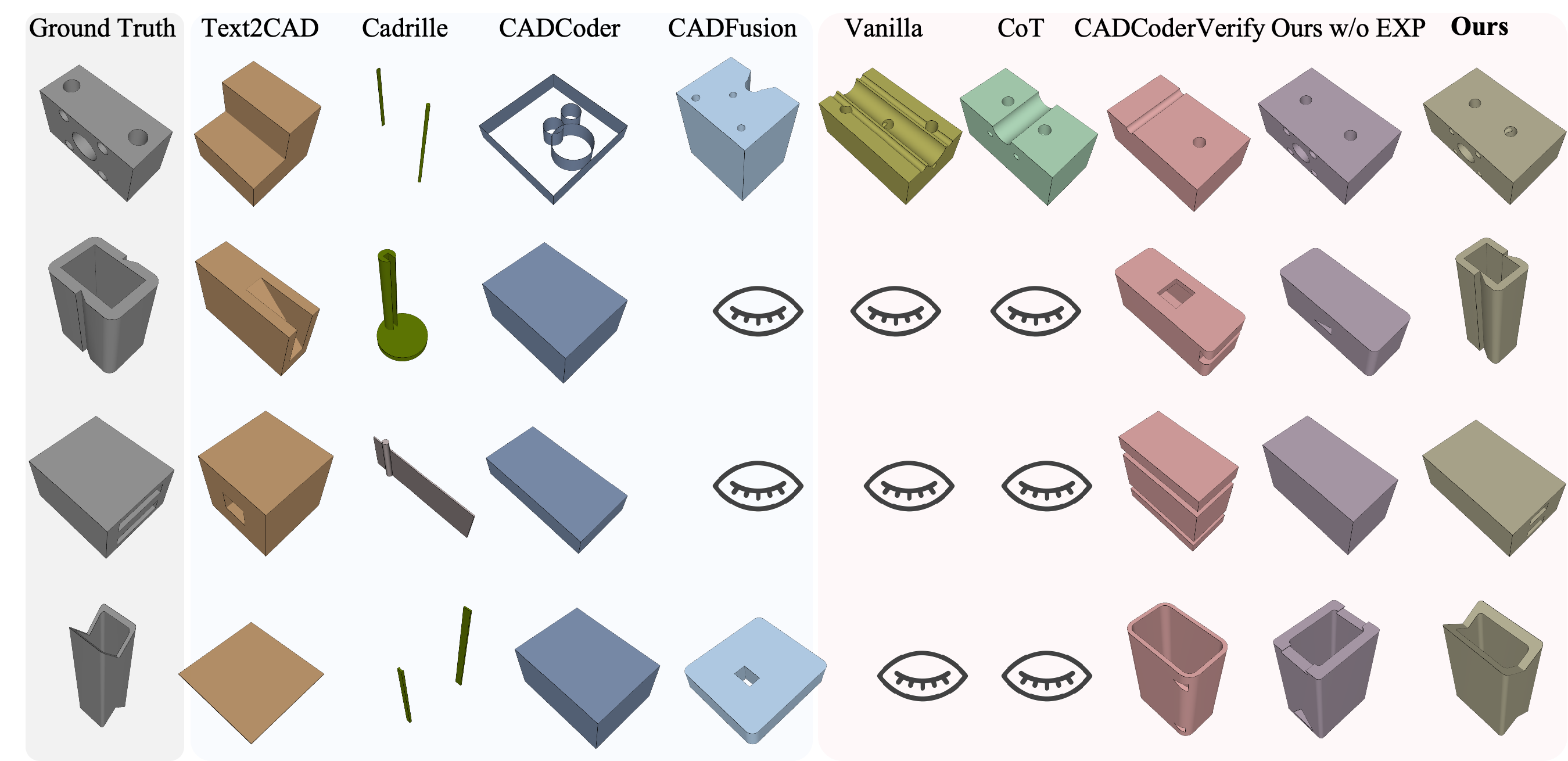}
    \caption{\textbf{Qualitative comparison on representative hard cases.}
Each row compares the ground-truth CAD model with outputs from different methods.
All LLM-based methods use Qwen3.5-27B as the backbone.
The closed-eye symbol denotes a rendering failure.
See Appendix~\ref{app:expconcad_cases} for input descriptions and generation traces.}
    \label{fig:qualitative_cases}
\end{figure*}

\subsection{Main Results}
\label{sec:main_results}

As shown in Table~\ref{tab:cadfusion-hard-main}, \textsc{ExpConCAD} achieves the highest $\mathrm{VLM}$ scores on both Qwen3.5-27B and GPT-5, indicating superior alignment with user descriptions. Notably, even without experience memory, \textsc{ExpConCAD} substantially outperforms existing methods, demonstrating that construction-aware spatial constraint reasoning can effectively recover construction structures and infer implicit spatial constraints from underspecified descriptions. Incorporating experience memory further improves $\mathrm{VLM}$ by 11.2\% on Qwen3.5-27B and 7.5\% on GPT-5, suggesting that reusable experience provides additional guidance for resolving missing constraints and better capturing user intent. Consistent qualitative evidence is shown in Figure~\ref{fig:qualitative_cases}, where \textsc{ExpConCAD} generates CAD models that more closely resemble the ground truth. For geometry-based metrics, \textsc{ExpConCAD} variants also achieve the strongest overall performance, although the gains from experience memory are less consistent. A possible explanation is that underspecified descriptions may correspond to multiple valid geometries. Consequently, overlap- and distance-based metrics can be sensitive to acceptable geometric variations even when the generated CAD model satisfies the intended design constraints. Overall, the results show that \textit{our spatial constraint reasoning is the primary driver of performance, while experience memory further improves constraint completion and description fidelity.}

\subsection{Ablation Study}
\label{sec:ablation}

\begin{figure}[t]
    \centering
    \includegraphics[width=\columnwidth]{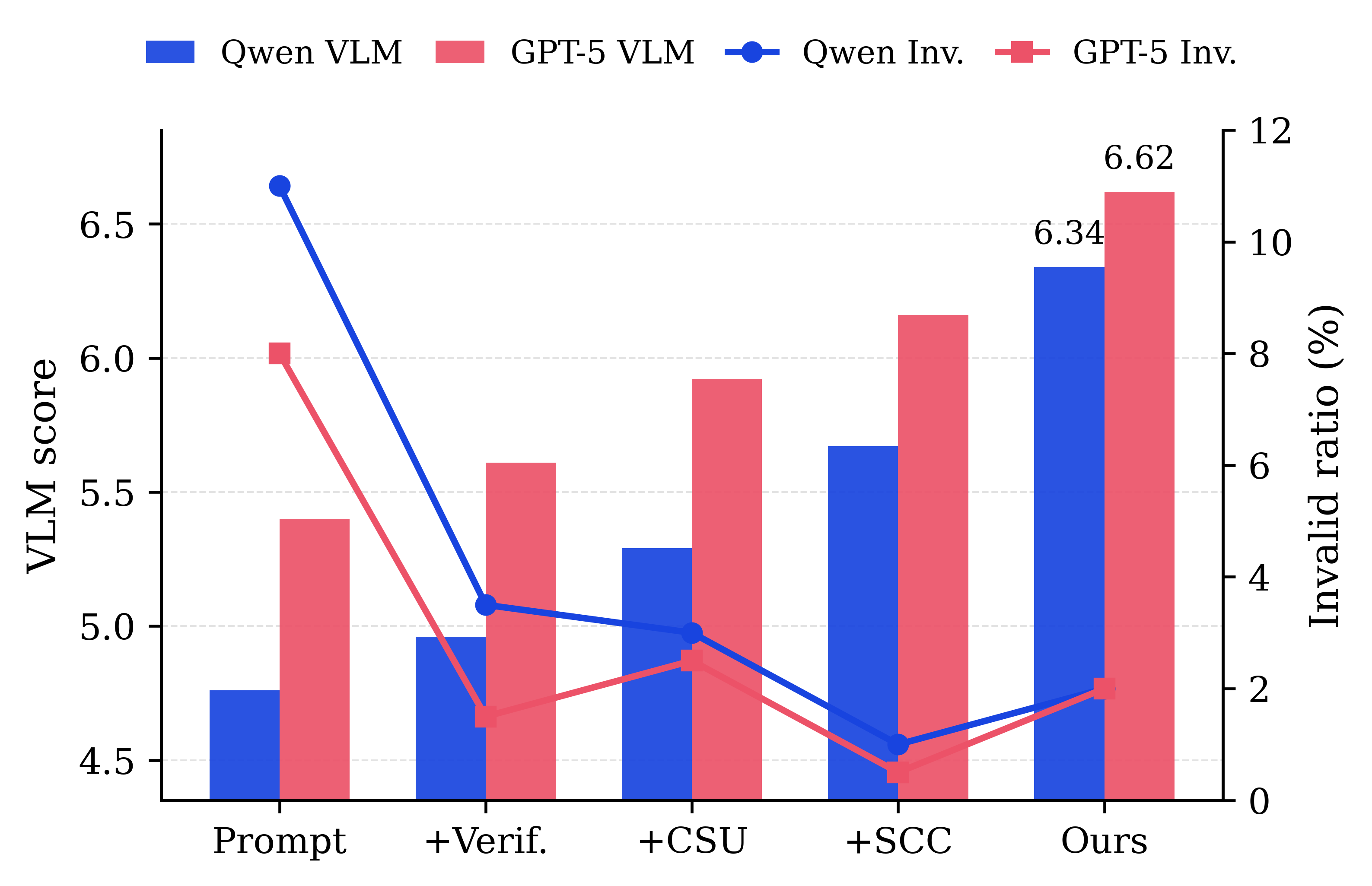}
    \caption{{Ablation study.}
    }
    \label{fig:ablation_cad}
\end{figure}

We conduct an ablation study to isolate the contribution of each major component in \textsc{ExpConCAD}.
As shown in Figure~\ref{fig:ablation_cad}, \textit{Prompt} denotes a strong task-specific prompt, \textit{+Verif.} adds execution-based verification and revision, \textit{+CSU} introduces Construction Structure Understanding, \textit{+SCC} adds Spatial Constraint Completion, and \textit{Ours} further uses experience-enhanced SCC.
Adding verification substantially reduces $\mathrm{Inv.}$.
Construction Structure Understanding and Spatial Constraint Completion then progressively improve $\mathrm{VLM}$ by identifying constraint scopes and completing missing spatial constraints.
The full framework achieves the best $\mathrm{VLM}$ score, showing that experience memory further improves constraint completion.
These results confirm that \textsc{ExpConCAD}'s gains come from combining executable-code verification, construction-aware spatial constraint reasoning, and experience-enhanced SCC.
Full metric results are provided in Appendix~\ref{app:ablation}, Table~\ref{tab:ablation-cadfusion-hard}.

\begin{figure}[t]
    \centering
    \includegraphics[width=0.95\columnwidth]{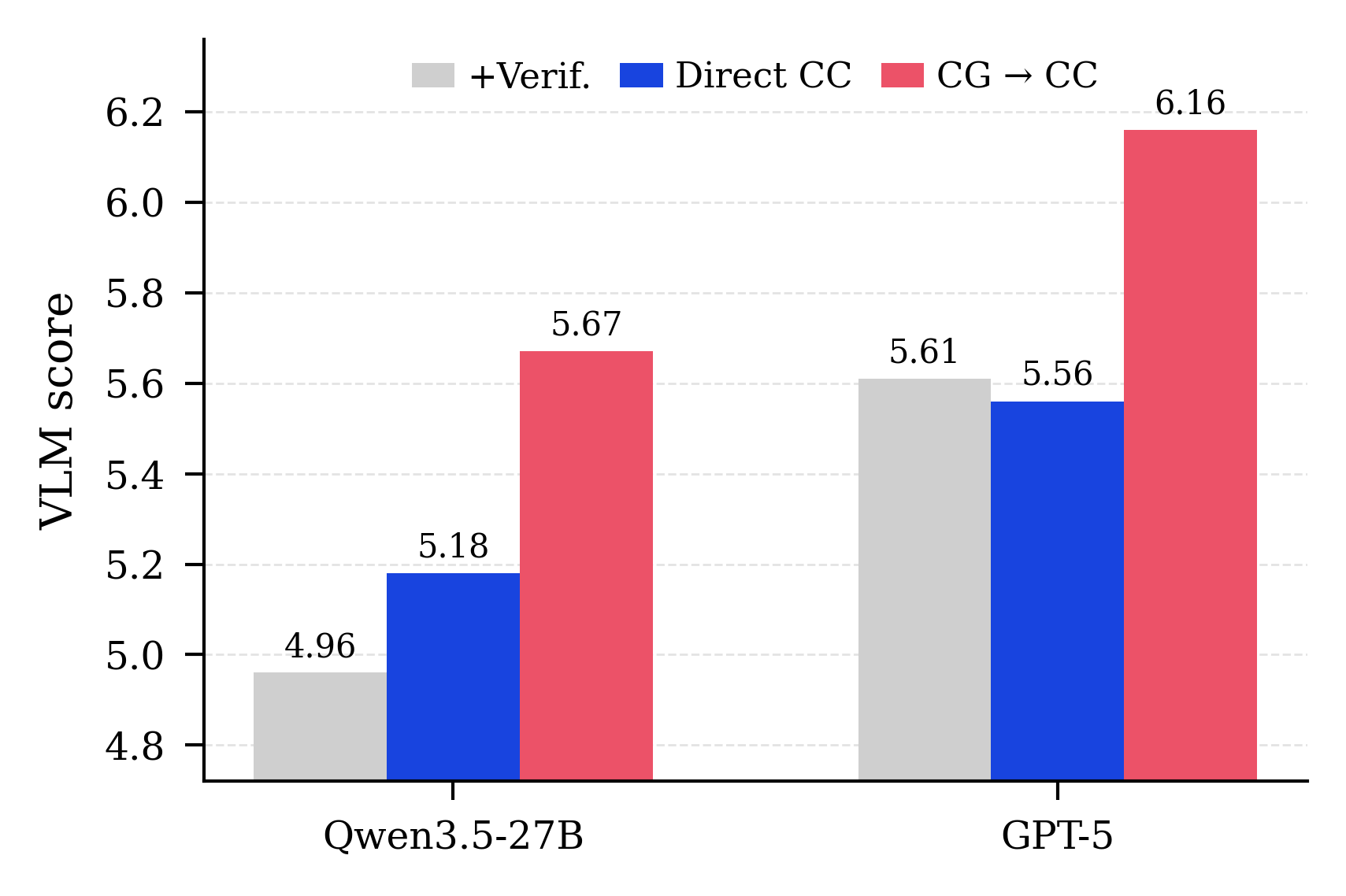}
    \caption{CSU as a prerequisite for SCC.}
    \label{fig:cg_cc_analysis}
\end{figure}

\subsection{Analysis of Key Design Choices}
\label{sec:analysis}
This section analyzes the effectiveness of \textsc{ExpConCAD} by addressing two questions: (1) Why is the CSU required before the SCC? and (2) How does experience memory facilitate the SCC? 
Detailed experimental settings are provided in Appendix~\ref{app:diagnostic_settings}.

\noindent\textbf{Does CSU provide a better basis for SCC?} 
To understand this, we perform analysis under three settings: a verification-only baseline, direct SCC from underspecified descriptions, and CSU followed by SCC. As shown in Figure~\ref{fig:cg_cc_analysis}\footnote{Full results are reported in Appendix~\ref{app:ablation}, Table~\ref{tab:ablation-cadfusion-hard}.}, direct SCC yields limited and unstable gains, whereas CSU+SCC consistently improves performance across both backbones. This suggests that underspecified descriptions alone can not specify the constraint scopes to be completed. By first recovering construction elements, operations, and constraint scopes, CSU provides SCC with explicit completion targets. Therefore, missing spatial constraints are more effectively inferred from the recovered construction structure than from the original description alone.



\begin{figure}[t]
    \centering
    \includegraphics[width=\columnwidth]{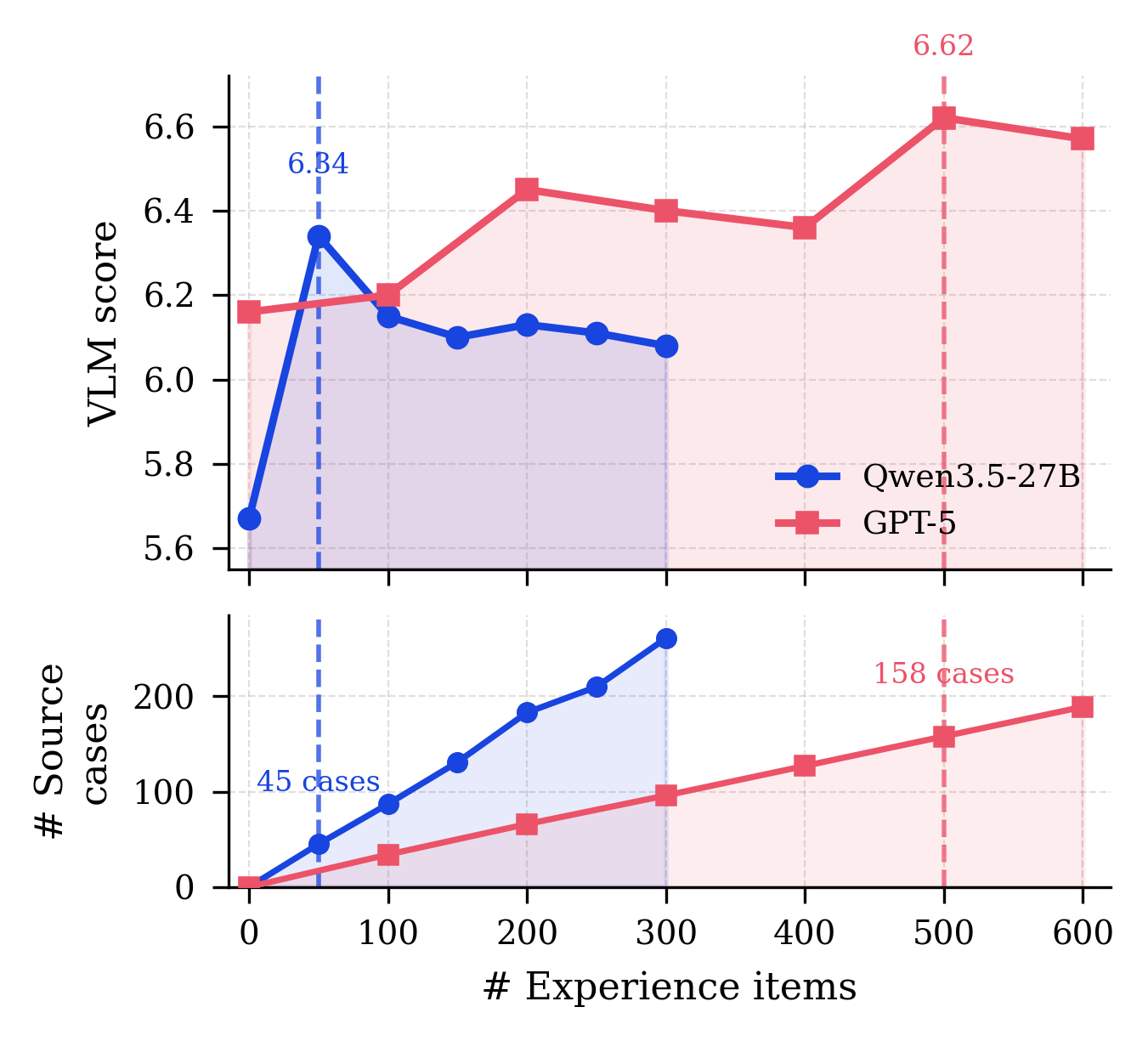}
    \caption{Effect of experience memory on SCC.}
    \label{fig:exp_size_analysis}
\end{figure}

\noindent\textbf{What makes experience useful for SCC?}
\label{sec:analysis_exp_size}
Figure~\ref{fig:exp_size_analysis} shows that experience improves SCC, but simply adding more experience does not always help.
For Qwen3.5-27B, performance peaks with 50 experience items and then saturates, suggesting that the added items are not always reusable for the current constraint-completion need.
A likely reason is that its experience often remains case-level: one item may mix several constraints from the original CAD case, making it hard to match a specific scope in a new case.
In contrast, GPT-5 benefits more from a larger memory.
Its experience is more often distilled into scope-specific patterns, such as cutter--base placement, cutter--cutter separation, or through-cut depth.
These patterns are easier to retrieve and reuse when a new case encounters a similar constraint scope.
Cases in Appendix~\ref{app:expconcad_cases} provide examples.
Overall, the key is not to store more cases, but to store reusable constraint-completion patterns at the right scope.

\subsection{Robustness and Experience Transfer}

\label{sec:generalization}
\begin{figure}[t]
    \centering
    \includegraphics[width=\columnwidth]{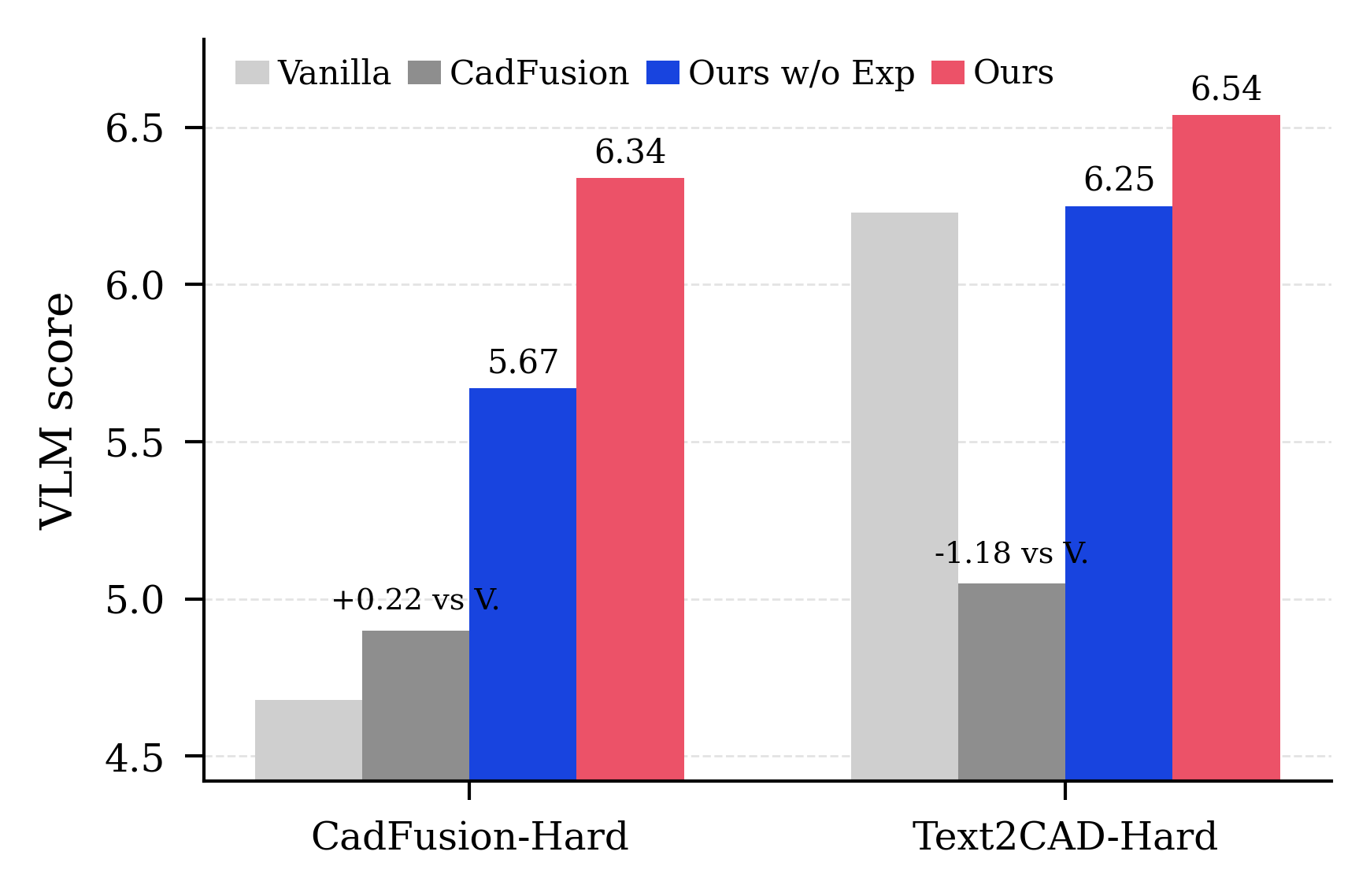}
    \caption{Robustness and experience transfer on unseen cases.}
    \label{fig:experience_generalize}
\end{figure}
We further evaluate the robustness of \textsc{ExpConCAD} and the transferability of its experience on unseen cases.
Using Qwen3.5-27B Vanilla as the reference baseline, we compare \textsc{ExpConCAD} with CadFusion on CadFusion-Hard and Text2CAD-Hard, where the latter follows a different shape distribution. (details in Appendix~\ref{app:generalization_settings}).
As shown in Figure~\ref{fig:experience_generalize}\footnote{Full results are provided in Appendix~\ref{app:text2cadhard-results}, Table~\ref{tab:text2cad-hard-main}.}, \textsc{ExpConCAD} improves over Vanilla on both datasets, with further gains from experience even on unseen {Text2CAD-Hard} cases.
By contrast, CadFusion improves on CadFusion-Hard but drops below Vanilla on Text2CAD-Hard.
These results suggest that construction-structure recovery and spatial constraint completion form a robust pipeline, while experience memory transfers through recurring constraint-completion patterns rather than source-dataset shape memorization.

\section{Conclusion}
In this work, we studied Text-to-CAD generation under underspecified descriptions, where missing spatial constraints hinder accurate CAD construction. To address this challenge, we proposed {\textsc{ExpConCAD}}, which combines construction-aware spatial constraint reasoning with an experience memory. 
More broadly, our work highlights a fundamental shift for CAD generation: from reconstructing geometry explicitly described by users to inferring the design knowledge that users leave unstated. We believe this ability to reason about missing design intent will be essential for building more human-centric CAD agents or copilots that collaborate naturally with human designers.


\clearpage
\section*{Limitations}

A key limitation of this work concerns experience management. Although the experience memory is constructed from prior CAD construction cases and used to guide spatial constraint completion, we do not study how it should be updated, deleted, merged, or validated over time. This is important because CAD constraint-completion patterns are often recurring and finite, so continuously expanding the memory may increase storage and retrieval costs without adding genuinely new patterns. In addition, multiple experiences may instantiate the same pattern but differ in quality or executability. Future work should develop memory management strategies to retain high-quality constraint-completion patterns, merge redundant experiences, and keep the memory compact and reliable.

\section*{Ethical Considerations}

This work aims to make CAD generation more accessible by enabling users to create CAD models
from natural-language descriptions, supporting rapid prototyping, education, and design
assistance. The study does not involve human subjects, sensitive personal data, or
privacy-sensitive information, and all source data come from open-source communities. Human
annotation and verification, where used, were conducted by internal CS Master's and PhD students
with CAD-related coursework or experience; participation was voluntary, and written instructions
were provided. Details are given in Appendix~\ref{app:vlm_2v}. We follow the ACL Code of Ethics,
and to the best of our knowledge, this study does not raise major concerns related to privacy,
bias, or safety.

\paragraph{Reproducibility}
To facilitate reproducibility, we will publicly release the source code on GitHub
(\url{https://anonymous.4open.science/r/ExpConCAD-B118}). Detailed setup and usage
instructions are provided in Appendix~\ref{app:expconcad_details} and in the repository
README. 

\paragraph{Artifact License and Usage Terms}
All code, data, and artifacts used or created in this work comply with their respective
licenses and usage terms. External code and datasets are open-source or publicly available,
and we respect their licenses, including proper attribution to the original creators.
The artifacts created by \textsc{ExpConCAD} are released under CC-BY 4.0 and are intended
for research purposes.

\paragraph{LLM Usage}
LLMs are used in this work as the backbone models of \textsc{ExpConCAD}, as described in
Appendix~\ref{app:method-details}. In addition, LLMs were used only for language polishing
during paper writing. All method design, experiments, analysis, and core ideas were developed
solely by the authors.

\bibliography{custom}

\appendix

\clearpage
\section{Details of \textsc{ExpConCAD}}
\label{app:expconcad_details}

\subsection{Construction-aware Spatial Constraint Reasoning}
In practice, the Construction-aware Spatial Constraint Reasoning consists of three stages. The first stage performs CAD construction structure understanding and generates queries for experience retrieval. The second stage conducts spatial constraint completion and produces an initial version of the CadQuery code. The final stage refines the generated code based on execution error messages.

\paragraph{CadQuery Function Guide}

\begin{figure*}[p]
    \centering
    \includegraphics[width=\textwidth]{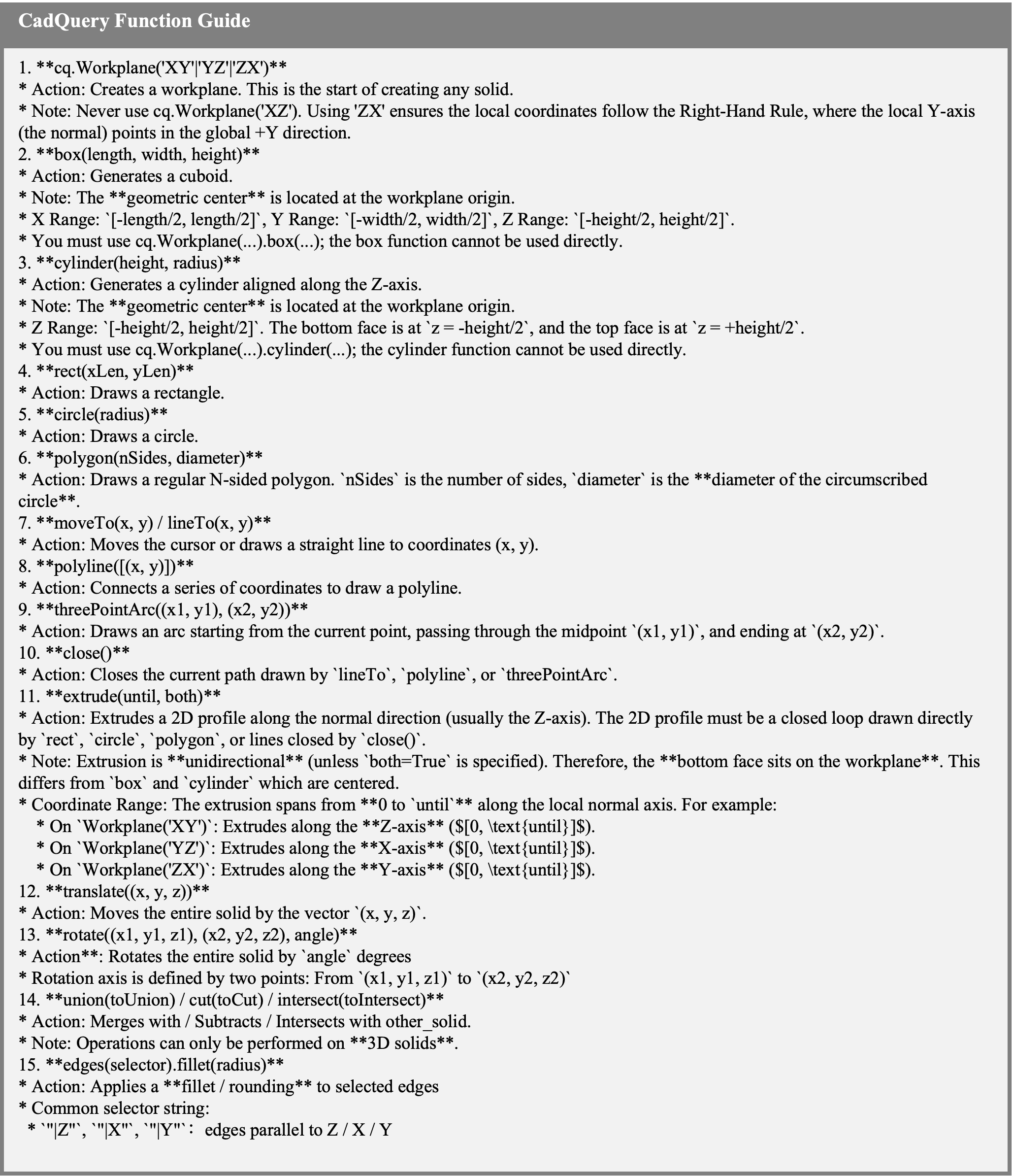}
\caption{
CadQuery function guide.
}
    \label{fig:prompt_func_guide}
\end{figure*}

Large language models commonly exhibit insufficient understanding of the CadQuery library, resulting in frequent syntax errors and semantic misuse of functions during code generation. To enhance model reliability, we introduce a curated \textit{Function Guide} that organizes the usage patterns and implementation details of essential CadQuery APIs. The guide is integrated into all three stages of \textsc{ExpConCAD} as supplementary knowledge, thereby improving generation accuracy and robustness. Figure~\ref{fig:prompt_func_guide} illustrates the constructed Function Guide.

\paragraph{Stage 1: Construction Structure Understanding and Query Generation}

\begin{figure*}[p]
    \centering
    \includegraphics[width=\textwidth]{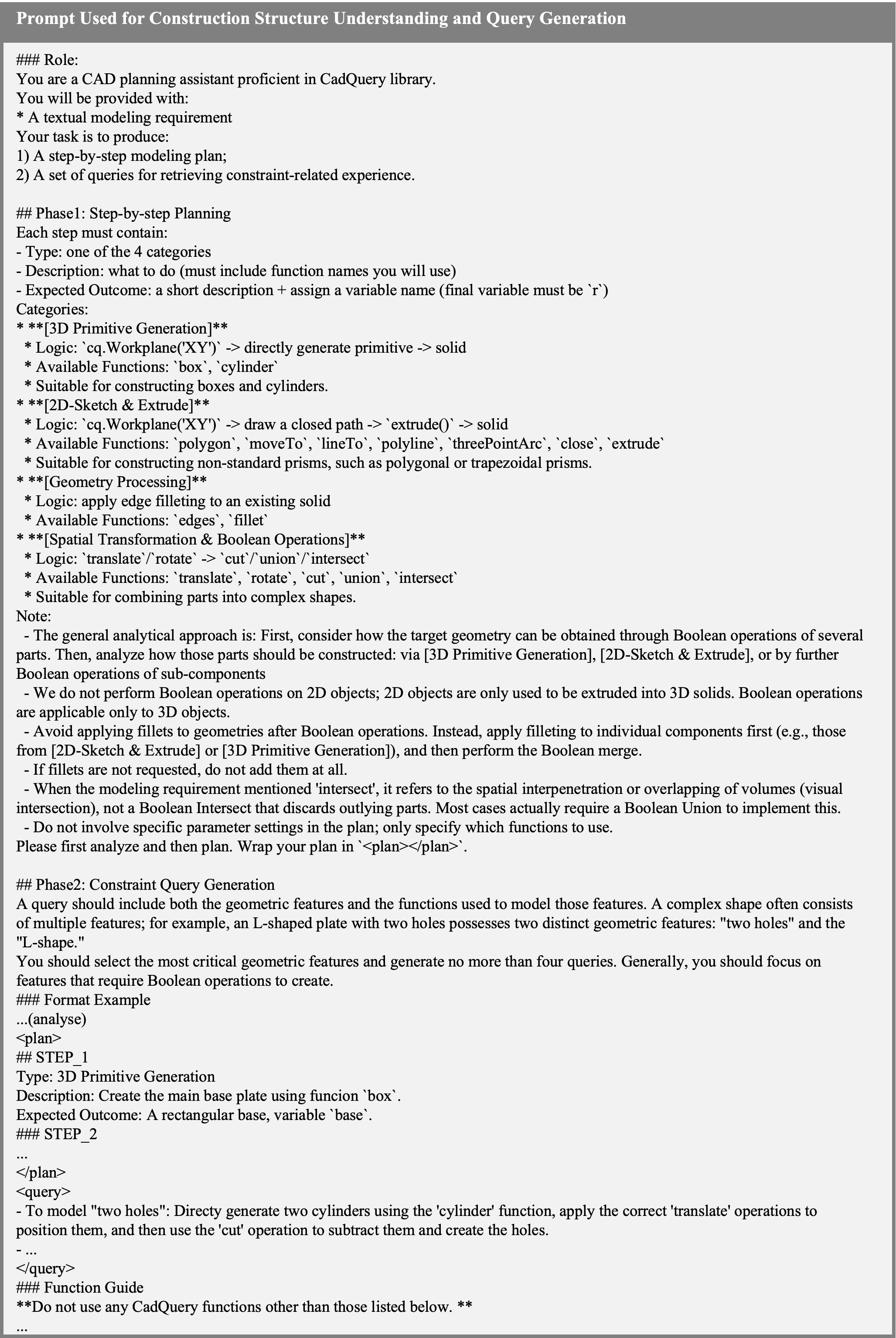}
\caption{
Prompt used for construction structure understanding and query generation.
}
    \label{fig:prompt_grounding_query}
\end{figure*}

The prompt used in Stage~1 is illustrated in Figure~\ref{fig:prompt_grounding_query}. To facilitate construction structure understanding, we categorize CAD modeling operations into four categories: \textit{3D Primitive Generation}, \textit{2D-Sketch \& Extrude}, \textit{Geometry Processing}, and \textit{Spatial Transformation \& Boolean Operation}. The first two categories are used to directly create CAD entities, the third category is used for geometric refinement operations such as filleting, and the last category is responsible for composing multiple entities.

During reasoning, the model first analyzes how the target geometry can be obtained through Boolean operations of several parts. Then, analyze how those parts should be constructed: via "3D Primitive Generation", "2D-Sketch \& Extrude", or by further Boolean operations of sub-components. Based on this reasoning process, the model identifies the CAD elements, modeling operations, and constraint scopes from the shape description, thereby providing a structured foundation for the subsequent spatial constraint completion.

This stage also generates queries for experience retrieval. Based on the construction structure, the model generates fine-grained queries containing CAD elements and the modeling operations between them, facilitating the retrieval of relevant modeling experiences.

\paragraph{Stage 2: Spatial Constraint Completion and Code Generation}

\begin{figure*}[p]
    \centering
    \includegraphics[width=\textwidth]{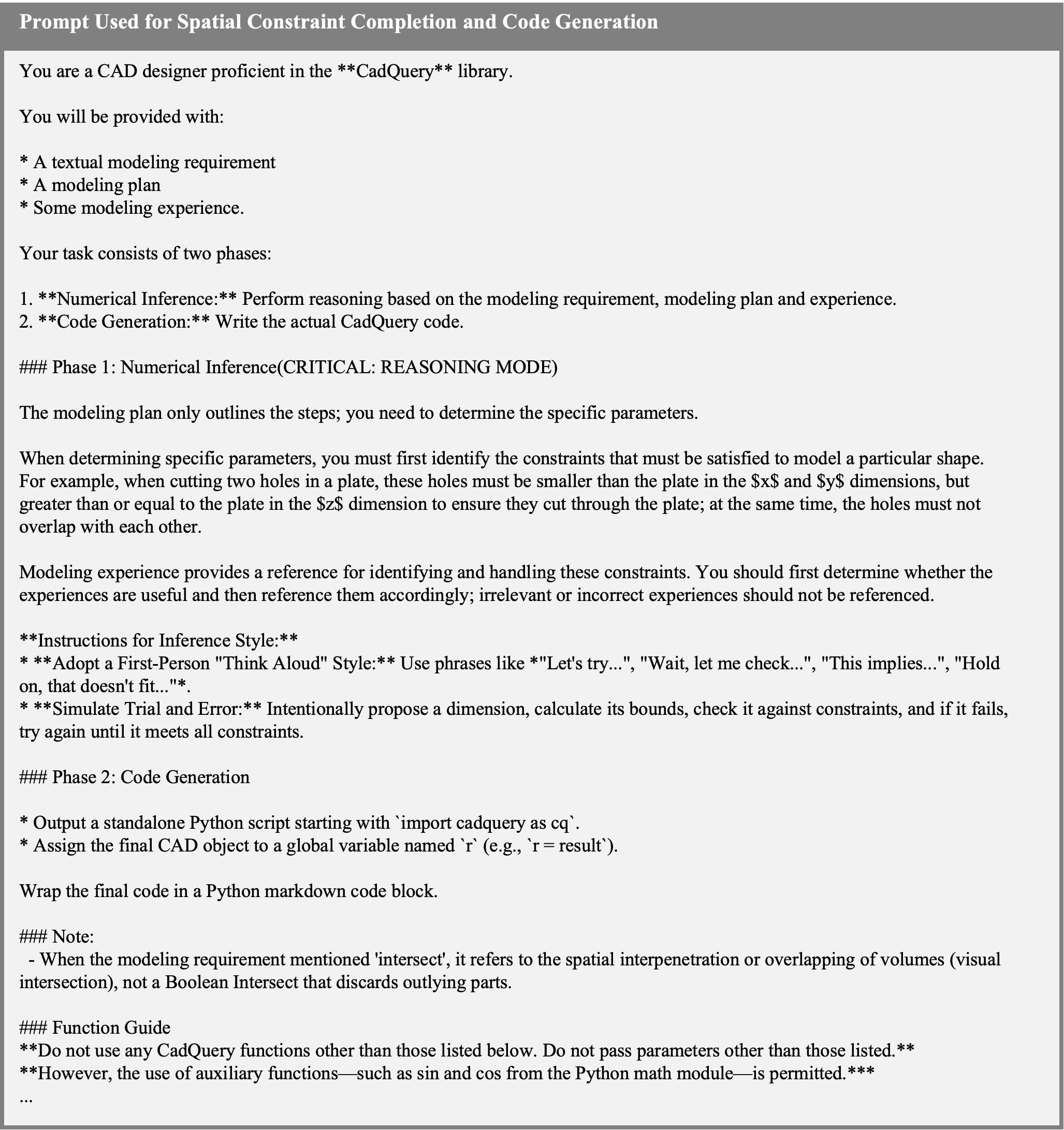}
\caption{Prompt used for spatial constraint completion and code generation.
}
    \label{fig:prompt_constraint_code}
\end{figure*}

The prompt used in Stage~2 is illustrated in Figure~\ref{fig:prompt_constraint_code}. Prior to this stage, relevant  experiences are retrieved using the query generated in Stage~1. We employ the BAAI General Embedding model (BGE), specifically \texttt{bge-large-en-v1.5}~\cite{bge_embedding}, to encode both the retrieval queries and the applicable constraint scope of the experiences into dense vector representations. Retrieval is then performed using cosine similarity in the embedding space, where the top-$k$ nearest experiences are selected as auxiliary context. In our implementation, the top three retrieved experiences are injected into the model context.

Given the original shape description, the construction structure, and the retrieved experiences, the model infers the implicit constraints and performs numerical reasoning to derive geometric parameter values that satisfy these constraints. Based on the construction structure and derived parameters, the model subsequently generates the corresponding CadQuery code.

\paragraph{Stage 3: Code Refinement}

\begin{figure*}[h]
    \centering
    \includegraphics[width=\textwidth]{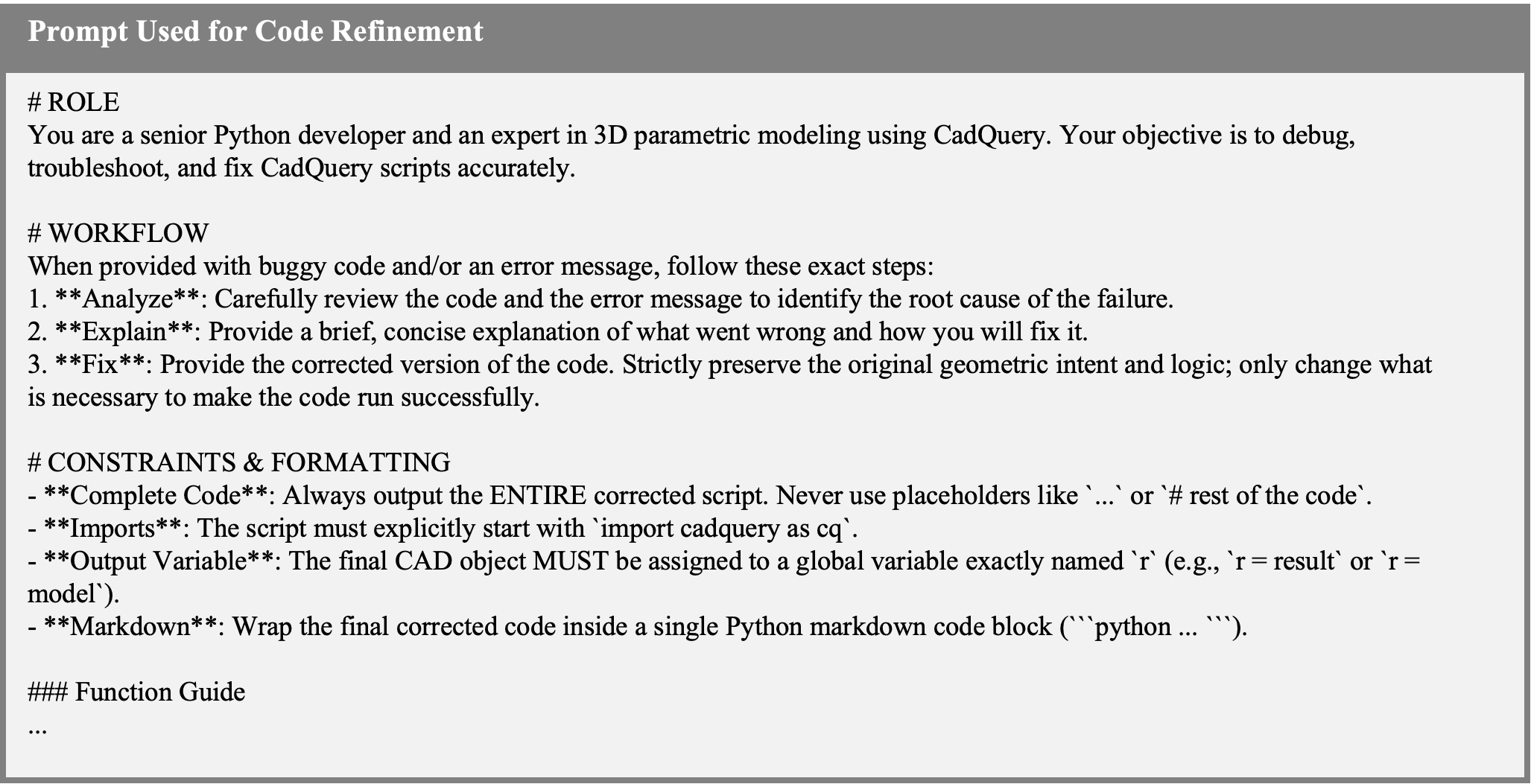}
\caption{
Prompt used for code refinement.
}
    \label{fig:prompt_refinement}
\end{figure*}

The prompt used in Stage~3 is illustrated in Figure~\ref{fig:prompt_refinement}. This stage focuses on correcting execution errors in the generated CadQuery code. When code execution fails, the executor returns the corresponding error messages to the model, which analyzes the failure causes and revises the implementation accordingly. The refinement stage only addresses syntax and implementation-level errors, without modifying the underlying modeling logic or numerical parameter configurations. The refinement process continues iteratively until the generated code executes successfully or a maximum of three refinement rounds is reached.

\subsection{Experience memory}
For each piece of data used for experience extraction, we first applied the construction-aware spatial constraint reasoning pipeline without experience enhancement to generate CadQuery programs. The programs were then executed and rendered into three-view images, which were subsequently evaluated by GPT-5 serving as the judge model. The judge assigned an initial score ranging from 0 to 10 and produced a corresponding diagnosis describing the major failure modes of the generated result. We only retained samples with initial scores lower than 7, which we regard as CAD generation failures, for subsequent experience distillation.

\begin{figure*}[p]
    \centering
    \includegraphics[width=\textwidth]{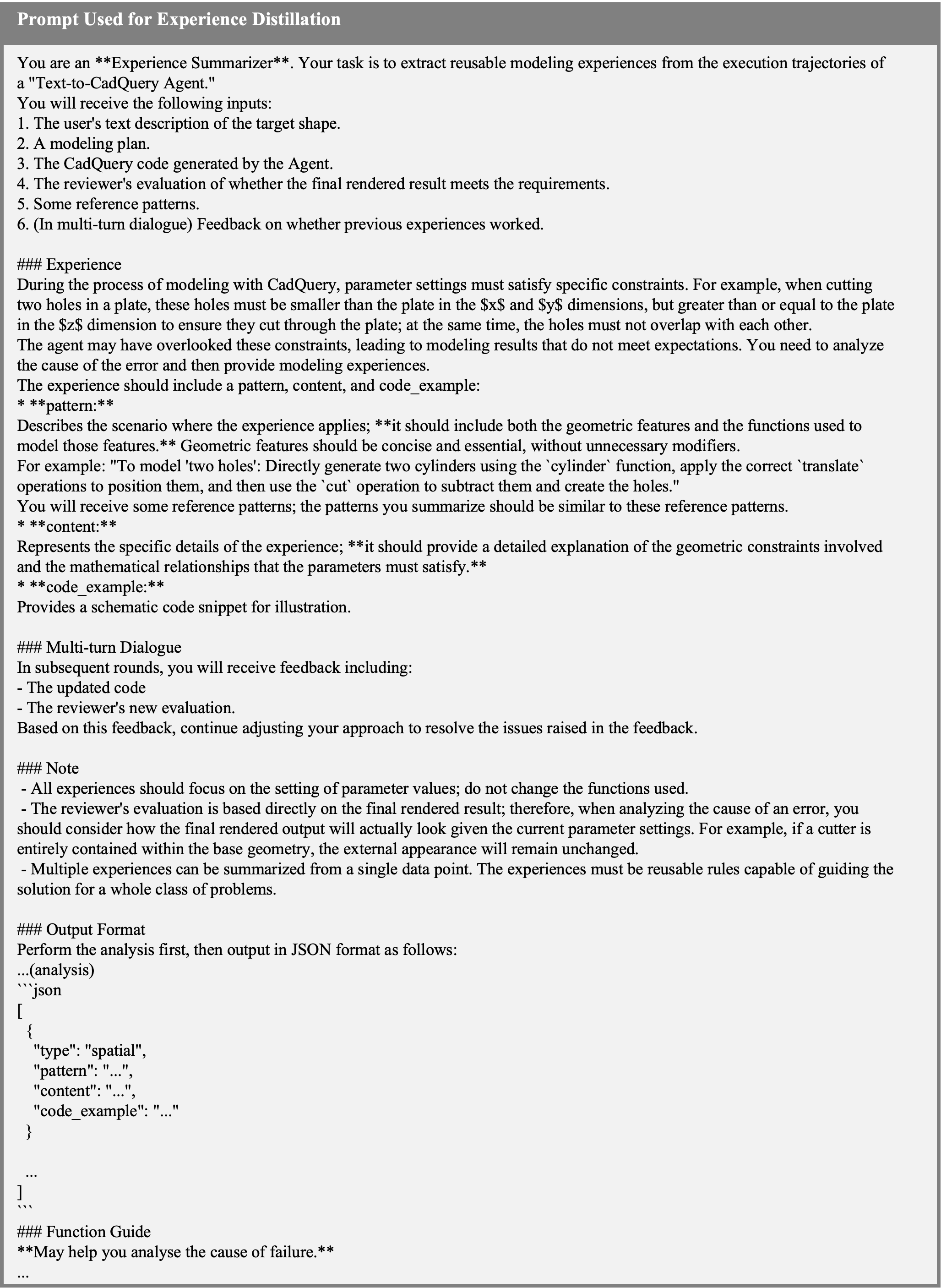}
\caption{
Prompt used for experience distillation.
}
    \label{fig:prompt_exp_distill}
\end{figure*}

The experience distillation process for each sample is formulated as a multi-turn iterative procedure with verification, ensuring that the retained experiences are validated through measurable improvements in generation quality. Specifically, given the shape description, construction structure, previously generated code, associated score and diagnostic feedback, we first analyze the root causes of the current CAD generation failure and distill a set of reusable modeling experiences. The prompt used for experience distillation is illustrated in Figure~\ref{fig:prompt_exp_distill}.

The distilled experiences are then injected back into the generation pipeline as additional contextual guidance. Conditioned on the combination of shape description, construction structure, and experiences, the model regenerates new code. The generated code is subsequently executed and rendered into three-view images and re-evaluated by GPT-5 to obtain updated scores and diagnoses.

If the score of regenerated result achieves an improvement of more than two points over the initial score, the corresponding experiences are considered verified and are stored in the final experience repository. Otherwise, the newly generated code, updated score, and revised diagnostic feedback are fed back into the experience distillation module for another round of reflection and refinement. This iterative process continues until either a sufficient score improvement is achieved or the maximum number of three retry iterations is reached.

\section{Case Studies of \textsc{ExpConCAD}}
\label{app:expconcad_cases}


Figure~\ref{fig:qualitative_cases} presents qualitative comparisons on four representative hard cases.
These examples cover common but challenging implicit constraints in vague-shape-level Text-to-CAD generation, including multi-face through holes, hollow structures, internal partitions, rounded corners, and repeated local cutouts.

In the first case, the input describes a mounting-block-like rectangular prism with two circular holes on the top face, a larger hole on the vertical face, and four smaller holes surrounding the larger one.
This requires inferring multi-face through-hole constraints and symmetric placement relations.
Most baselines fail to preserve the complete hole topology or the spatial arrangement of holes on different faces.
Our method better recovers the intended multi-face structure and the relative placement among holes.

In the second case, the target shape is an elongated hollow rectangular prism with rounded corners and triangular cutouts on two sides.
Existing methods often collapse the hollow structure, miss the side cutouts, or generate unrelated local geometry.
Our method better preserves the hollow body and recovers the side cutout interactions.

In the third case, the shape contains two parallel internal horizontal bars that create two rectangular voids.
This requires inferring internal partition constraints and symmetry from a vague description.
Baselines tend to generate a solid block, miss the internal bars, or produce incorrect separations.
Our method better captures the parallel bars and the resulting void structure.

In the fourth case, the input describes a tall hollow rectangular prism with open top and bottom, rounded corners, and identical triangular cuts near the top and bottom.
This case requires combining hollow-shell constraints, rounded-corner geometry, and repeated cutout constraints.
Our method better preserves these coupled construction constraints, producing a CAD model closer to the ground truth.

Overall, the qualitative results show that \textsc{ExpConCAD} is especially effective when the intended shape depends on implicit spatial constraints rather than explicit dimensions.
Together with the quantitative results, these cases demonstrate that CAD-context grounding, implicit constraint completion, and reusable experience jointly improve vague-shape-level Text-to-CAD generation.

\begin{figure*}[p]
    \centering
    \includegraphics[width=\textwidth]{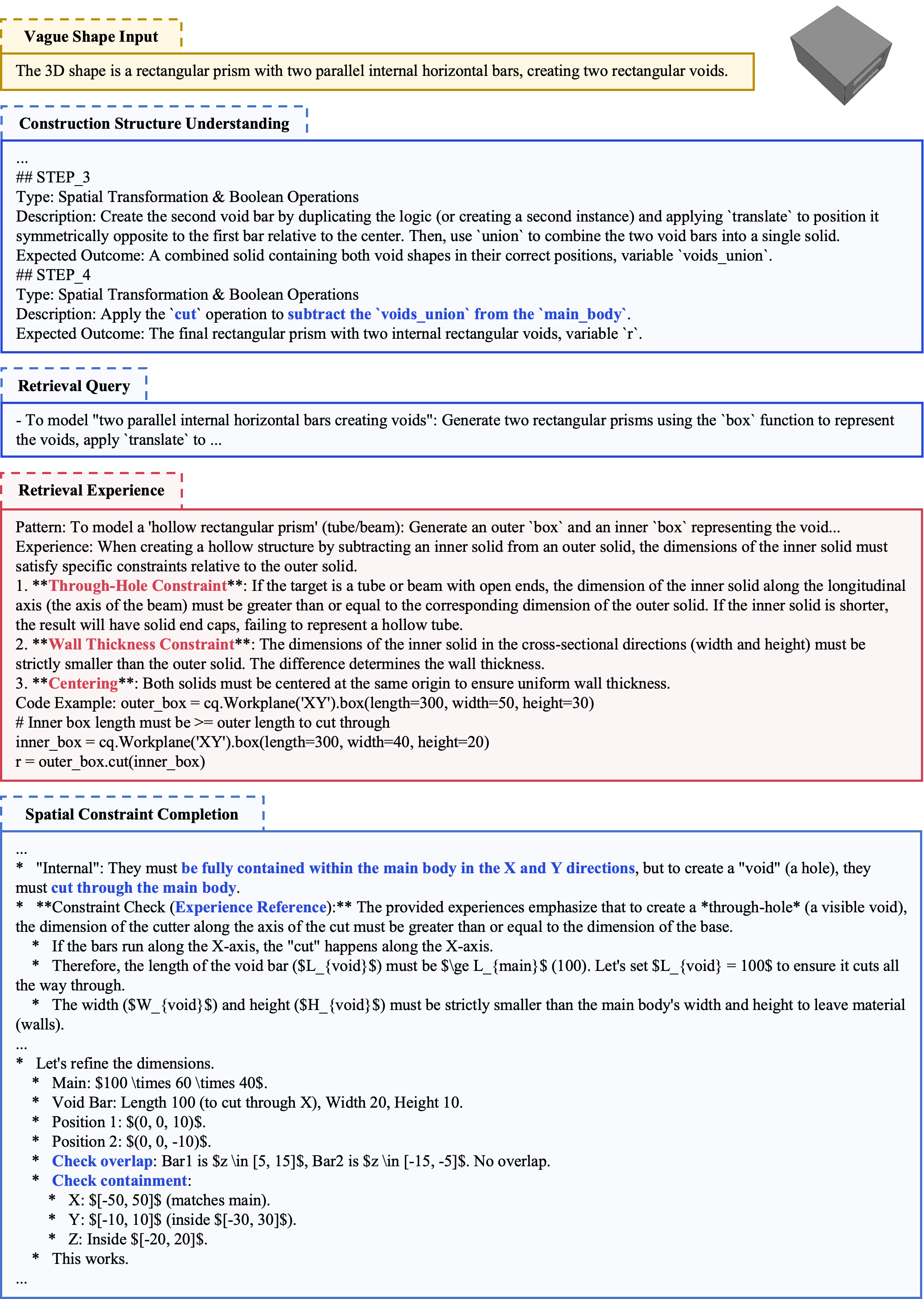}
\caption{
The full generation process for the third case in Figure~\ref{fig:qualitative_cases}.
}
    \label{fig:full_case}
\end{figure*}

To provide a more concrete understanding of how \textsc{ExpConCAD} performs implicit spatial constraint completion, we present the full generation process for the third case, as illustrated in Figure~\ref{fig:full_case}. Faced with the vague shape input “The 3D shape is a rectangular prism with two parallel internal horizontal bars, creating two rectangular voids,” the model first performs construction structure understanding, identifying the interactions between the CAD elements “voids\_union” and “main\_body.” This establishes a solid structural basis for subsequent spatial constraint completion. 

A query is then constructed and relevant experiences are retrieved. These experiences provide additional guidance on the constraints that should be considered for similar geometric patterns. For example, the "Through-Hole Constraint" is directly applicable to the current modeling task. During the implicit constraint completion process, the model leverages the construction structure together with the retrieved experiences to identify a set of constraints, such as ensuring that the voids are fully contained within the main body along the X and Y directions and that they cut through the main body. Based on these constraints, the model subsequently infers parameter configurations that satisfy all constraints.

Compared with other methods, our approach explicitly establishes interactions between CAD elements through construction structure understanding, making it easier to determine which inter-element constraints should be inferred. Furthermore, the introduction of the experience mechanism helps the model recognize additional necessary constraints. As shown in Figure~\ref{fig:qualitative_cases}, “Ours w/o EXP” ignores the constraint that the voids must be at least as large as the main body along a certain dimension in order to cut through it, resulting in voids that are entirely enclosed within the main body. By incorporating experiences, the model becomes aware of this constraint and successfully generates the correct structure.



\section{Implementation Details}\label{app:method-details}

\subsection{Benchmark Information}\label{app:benchmark}


We describe the benchmarks used in our experiments, including Text2CAD, CADFusion, and our constructed CADFusion-Hard. We report the dataset statistics and task settings for each benchmark, and further detail the construction procedure of CADFusion-Hard.

\paragraph{Text2CAD.}
Text2CAD~\cite{khan2024text2cad} is built upon the DeepCAD~\cite{2021DeepCAD} dataset and augments CAD models with multi-level natural language annotations generated by LLMs and VLMs. The benchmark defines four prompt levels, ranging from vague-shape-level descriptions (L0) to modeling-step-level descriptions (L3). CAD models are represented as sketch-and-extrude token sequences\cite{willis2021fusion}. In total, the benchmark contains approximately 600K training pairs and 32K validation/test pairs derived from around 150K CAD models.

\paragraph{CADFusion.}
CADFusion\cite{wang2025cadfusion} is also based on DeepCAD and adopts sketch-and-extrude CAD representations. Unlike most Text-to-CAD datasets that emphasize modeling-step-level descriptions, CADFusion mainly focuses on vague-shape-level descriptions. The captions are first generated from rendered CAD images using VLMs, and are subsequently refined and verified by human annotators to improve annotation quality. The resulting benchmark contains approximately 20K text-CAD pairs, among which 952 samples are used for testing.


\paragraph{Implicit-constraint hard subset construction.}
To evaluate models on genuinely challenging shape descriptions with implicit spatial constraints, we construct hard subsets from both CADFusion and Text2CAD.
For each benchmark, we first run a pool of standard baseline methods, including vanilla prompting and CoT prompting, for three independent runs, and compute the average VLM score for each sample across all method-run combinations.
We use this average score as an automatic difficulty proxy and select the 500 samples with the lowest scores as candidate hard cases.
However, low VLM scores may also result from noisy descriptions, low-quality samples, or ambiguous ground truth rather than genuine spatial reasoning difficulty.
Therefore, we further conduct human difficulty assessment.
Three CAD-experienced annotators independently rate each candidate from 0 to 10, producing a human difficulty score by averaging their ratings.
Annotators are instructed to assign high scores to valid but underspecified shape descriptions whose difficulty mainly comes from missing implicit spatial constraints, such as constraint scopes, feature placements, alignments, containment, spacing, or cut-through relations, and to assign low scores to noisy, unclear, or low-quality samples.
We then select the 200 samples with the highest human difficulty scores from each benchmark to form \uline{CADFusion-Hard} and \uline{Text2CAD-Hard}, respectively.
This procedure filters out low-quality cases while retaining samples whose difficulty mainly comes from missing spatial constraints.

\subsection{Experience Construction}
\label{app:construct_exp}


In the practical implementation of the experience distillation process, We randomly sampled 1,500 instances from the CADFusion\cite{wang2025cadfusion} training set for experience construction.

\subsection{Evaluation Metrics}\label{app:metrics}
We evaluate generated outputs from three complementary perspectives: \emph{(i)} text--shape consistency, \emph{(ii)} geometric fidelity to the reference shape, and \emph{(iii)} executability of the generated \textsc{CadQuery} program.

\paragraph{VLM score.}
\label{app:vlm_2v}
We use VLM score to evaluate the semantic consistency between the input description and the generated CAD model.



Traditional VLM-based evaluation typically renders a 3D shape into orthographic three-view images before scoring, referred to as \textbf{VLM$_{3v}$}. However, the visual information provided by three-view renderings is inherently limited. We propose \textbf{VLM$_{2v}$}. Given a generated shape, we render it from two informative viewpoints and query a vision-language model to assess whether the rendered object matches the input text description. Let $x$ denote the text prompt, $I^{(1)}$ and $I^{(2)}$ the two rendered views, and $f_{\mathrm{vlm}}(x, I^{(1)}, I^{(2)})$ the scalar score produced by the VLM. The metric is defined as
\begin{equation}
\mathrm{VLM}_{2v}(x, \hat{S}) = f_{\mathrm{vlm}}\bigl(x, I^{(1)}(\hat{S}), I^{(2)}(\hat{S})\bigr),
\end{equation}
where $\hat{S}$ denotes the generated 3D shape.

\begin{figure}[t]
    \centering
    \includegraphics[width=\columnwidth]{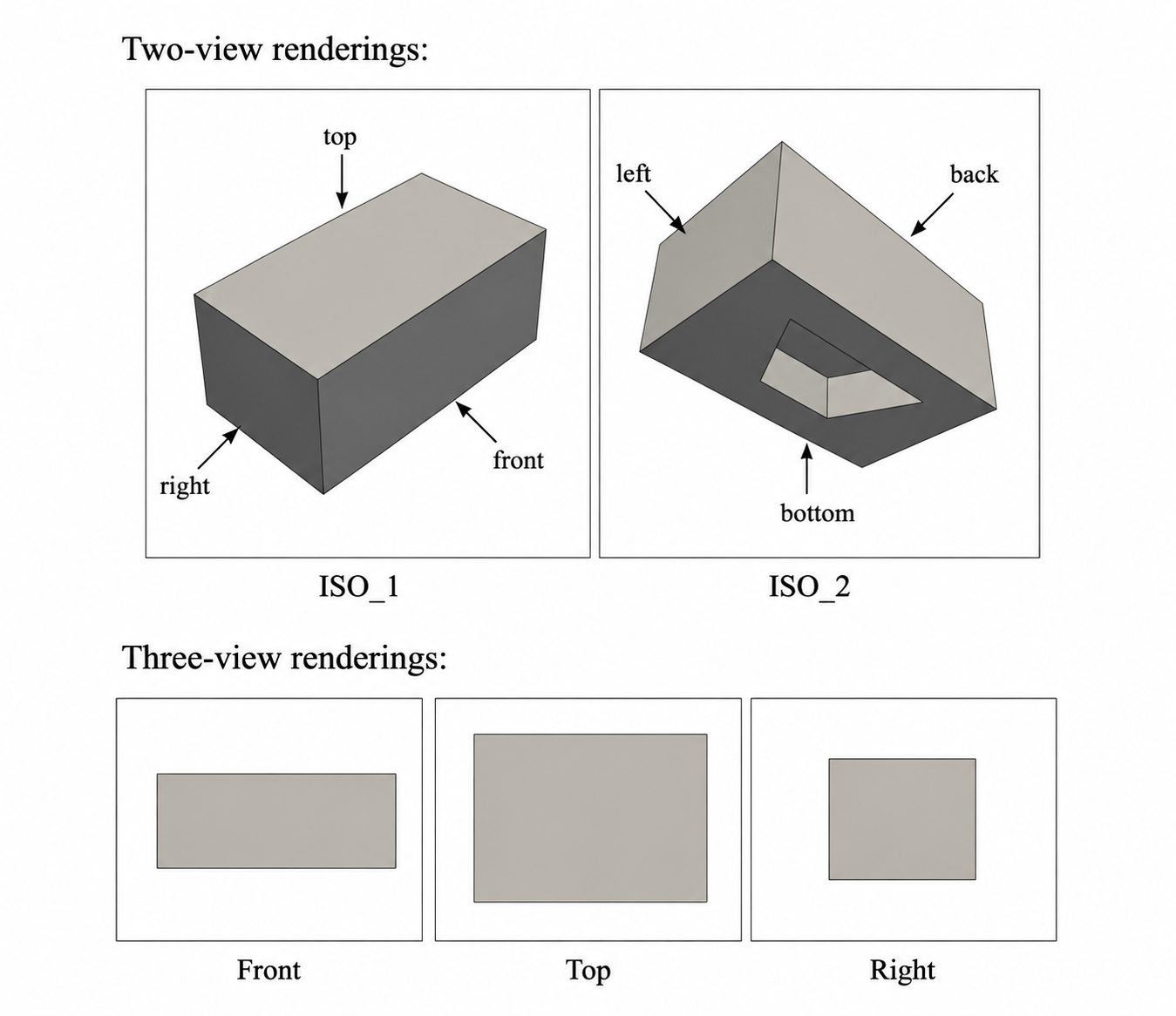}
\caption{
Comparison of Two-view and Three-view Renderings.
}
    \label{fig:2-view}
\end{figure}

Unlike conventional orthographic three-view renderings (\textit{i.e.}, front, top, and right), we adopt two complementary diagonal perspective views. This design is motivated by viewpoint selection principles in computer vision and graphics, where informative viewpoints are expected to maximize visible surface information while minimizing occlusion redundancy~\cite{vazquez2001viewpoint}.
Although orthographic projections are well suited for precise engineering specification, they are often suboptimal for perceptual completeness and semantic understanding. In particular, axis-aligned views frequently suffer from severe self-occlusion, causing substantial surface regions to become hidden or collapse into edge-aligned projections. By contrast, diagonal viewpoints reveal multiple principal faces simultaneously, substantially increasing effective surface visibility compared to standard three-view renderings, as shown in Figure~\ref{fig:2-view}, where features on the bottom face—largely obscured in the three-view renderings—become clearly visible, highlighting the advantage of the two-view perspective in conveying complete surface information.

Following this principle, each mesh is first normalized into a unit bounding box and centered at the origin. We then place two virtual cameras along approximately opposite diagonal directions relative to the object center, corresponding to the positive and negative octants:\quad(1,1,1)\quad \text{and}\quad (-1,-1,-1).
Both cameras are oriented toward the object centroid, while their focal distances are adaptively adjusted according to object scale to ensure complete coverage. This configuration produces two complementary isometric-like views that jointly capture the front/back, left/right, and top/bottom geometric structures.

\begin{figure*}[p]
    \centering
    \includegraphics[width=\textwidth]{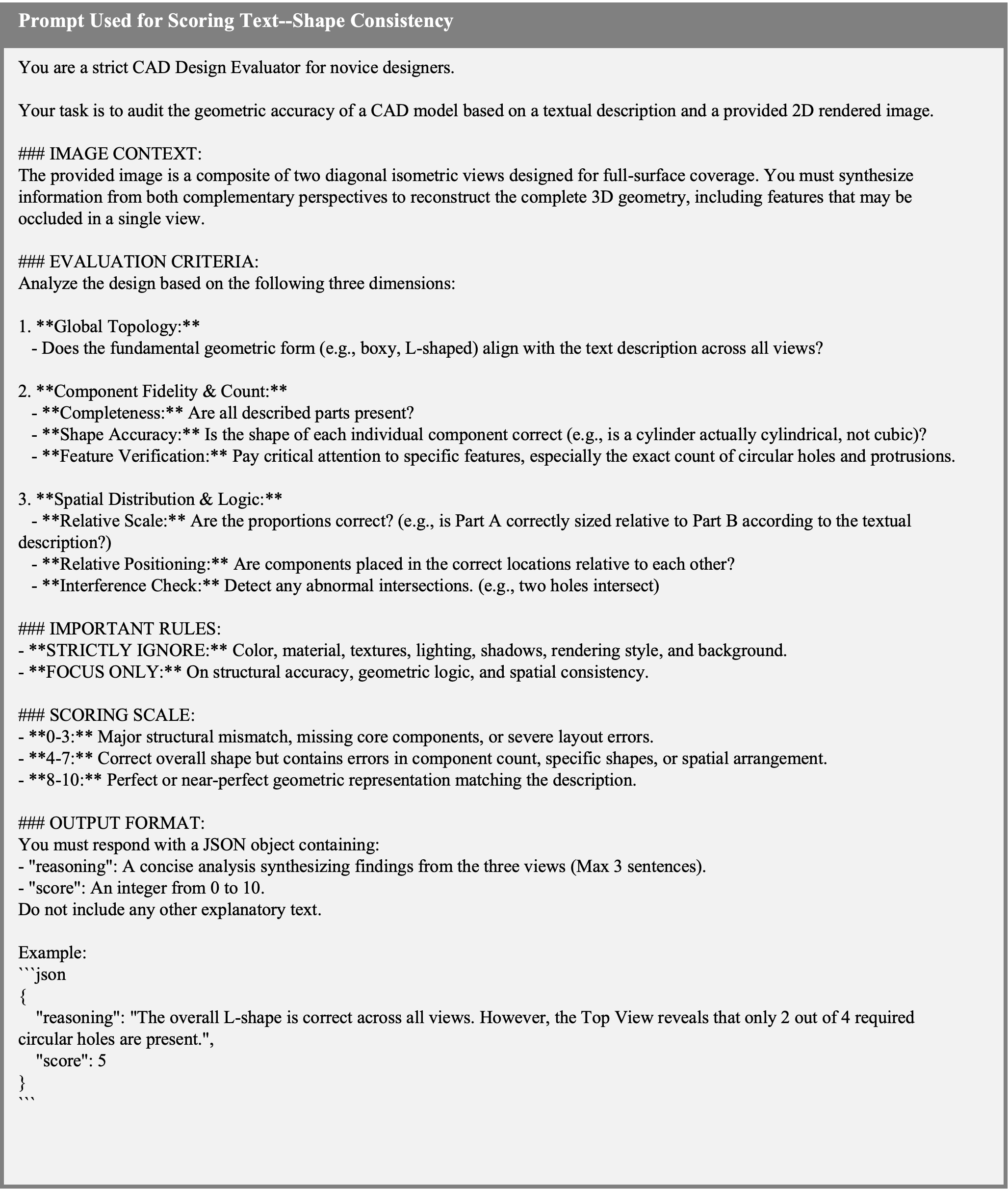}
\caption{
Prompt used for scoring text--shape consistency.
}
    \label{fig:prompt_vlm_eval}
\end{figure*}

The exact evaluation prompt used for VLM-based scoring is shown in Figure~\ref{fig:prompt_vlm_eval}. Following CADFusion~\cite{wang2025cadfusion}, the prompt is designed around three complementary evaluation dimensions: \textit{(1) global topology}, which measures whether the overall geometric structure aligns with the textual description; \textit{(2) component fidelity and count}, which verifies the presence, geometric correctness, and exact number of key components (e.g., holes and protrusions); and \textit{(3) spatial distribution}, which evaluates whether the relative scale and placement of components are correct, as well as whether abnormal intersections or other structural inconsistencies are present. Together, these dimensions encourage the VLM to perform holistic geometric reasoning rather than relying on superficial visual cues. To further mitigate rendering-induced bias, the prompt explicitly instructs the VLM to ignore color, texture, and other non-geometric cues,  focusing solely on structural and spatial correctness. The VLM is required to return both a textual rationale and a single integer score ranging from 0 to 10, where higher scores indicate stronger text--shape consistency.

To further evaluate the alignment between our evaluation metric and human preference, we conduct a human correlation study. We randomly sample 50 text--shape pairs and recruit five volunteer annotators with proficient CAD modeling experience to independently assess text--shape consistency on a 10-point scale. During evaluation, annotators are allowed to freely inspect each 3D model from arbitrary viewpoints using interactive visualization software. To ensure consistency between human assessment and the automatic evaluation framework, annotators are instructed to score each sample according to the same three evaluation dimensions used in the VLM prompt(Figure~\ref{fig:prompt_vlm_eval}). The final human score for each sample is obtained by averaging the ratings across annotators.We then compute the Spearman rank correlation coefficient to measure the ranking consistency between human judgments and two automatic metrics, namely \textbf{VLM$_{2v}$} and \textbf{VLM$_{3v}$}. Given paired samples $\{(y_i, \hat{y}_i)\}_{i=1}^{N}$, where $y_i$ and $\hat{y}_i$ denote human and model scores, respectively, the Spearman correlation is defined as
\begin{equation}
\rho = 1 - \frac{6 \sum_{i=1}^{N} d_i^2}{N(N^2 - 1)},
\end{equation}
where $d_i$ is the difference between the ranks of $y_i$ and $\hat{y}_i$.

The results, shown in Table~\ref{tab:human_correlation}, indicate that \textbf{VLM$_{2v}$} achieves higher correlation with human judgments than \textbf{VLM$_{3v}$}, suggesting that the proposed two-view rendering strategy better aligns with human perceptual preferences for text--shape consistency evaluation.

\begin{table}[t]
\centering
\small
\begin{tabular}{lc}
\hline
\textbf{Method} & \textbf{Spearman $\rho$} \\
\hline
\textbf{VLM$_{3v}$}& 0.71 \\
\textbf{VLM$_{2v}$} & \textbf{0.84} \\
\hline
\end{tabular}
\caption{Spearman correlation between automatic metrics and human judgments on text--shape consistency. Higher values indicate better alignment with human preference.}
\label{tab:human_correlation}
\end{table}

\paragraph{Rotation Invariant IoU.}
The Intersection-over-Union (IoU) is a common metric for evaluating the volumetric overlap between a predicted voxel grid $V_{\text{pred}}$ and a ground-truth voxel grid $V_{\text{gt}}$\cite{2025cadrille}, defined as
\begin{equation}
    \text{IoU} = \frac{|V_{\text{pred}} \cap V_{\text{gt}}|}{|V_{\text{pred}} \cup V_{\text{gt}}|}.
\end{equation}

In our text-to-CAD generation task, vague-shape-level descriptions often specify only the rough geometry of an object without constraining its orientation. As a result, the predicted object may be rotated relative to the ground-truth canonical orientation, which can lead to an unfair evaluation: for example, a cuboid rotated by $90^\circ$ around any principal axis remains semantically consistent with the text prompt, yet its voxel-wise overlap with the ground-truth in the fixed canonical orientation would be substantially reduced. To obtain an objective assessment that is insensitive to such pose variations, we adopt a rotation-invariant IoU (r-IoU) metric.

Specifically, we observe that in practice, most orientation misalignments occur as discrete rotations by multiples of 90$^\circ$ along the principal $x, y,$ or $z$ axes. To account for these discrepancies, we generate 24 discrete rotations of the ground-truth voxel grid for each object. These rotations correspond to the 24 elements of the cube rotation group \cite{conway2016symmetries}. Since this group captures every possible combination of 90$^\circ$ rotations, it effectively covers all major canonical orientations a typical CAD object may assume. Given the predicted voxel grid $V_{\text{pred}}$ and the set of rotated ground-truth voxel grids $\{V^{(r)}_{\text{gt}}\}_{r=1}^{24}$, the rotation-invariant IoU is defined as:
\begin{equation}
    \text{r-IoU} = \max_{r \in \{1,\dots,24\}} \frac{|V_{\text{pred}} \cap V^{(r)}_{\text{gt}}|}{|V_{\text{pred}} \cup V^{(r)}_{\text{gt}}|}.
\end{equation}

This approach ensures that evaluation focuses on shape correctness rather than penalizing orientation misalignment.

\paragraph{Chamfer Distance.}
The Chamfer Distance (CD) is another commonly used metric for evaluating geometric similarity between 3D shapes\cite{khan2024text2cad,2025cadrille,wang2025cadfusion}. It measures geometric similarity between two point sets sampled from mesh surfaces. Given a predicted point cloud $P$ and a ground-truth point cloud $G$, CD is defined as
\begin{equation}
\begin{split}
\text{CD}(P, G) = & \frac{1}{|P|} \sum_{p \in P} \min_{g \in G} \|p - g\|_2^2 \\
& + \frac{1}{|G|} \sum_{g \in G} \min_{p \in P} \|g - p\|_2^2.
\end{split}
\end{equation}

This metric captures how closely the predicted surface approximates the ground-truth geometry. We uniformly sample 2000 points from each mesh surface for CD computation. Following common practice\cite{khan2024text2cad}, all reported CD values in the paper are multiplied by $1000$ for readability.

\paragraph{Rotation Invariant Chamfer Distance.}
Similar to rotation-invariant IoU, we adopt the Rotation Invariant Chamfer Distance (r-CD) to account for unknown orientation. Specifically, we evaluate the Chamfer Distance across a discrete set of 24 pre-rotated ground-truth point clouds $\{G^{(r)}\}_{r=1}^{24}$ and take the minimum:
\begin{equation}
    \text{rCD} = \min_{r \in \{1, \dots, 24\}} \text{CD}(P, G^{(r)})
\end{equation}

\paragraph{Invalid ratio.}
A generated program is useful only if it can be successfully executed to produce a valid 3D model. Programs may fail due to syntax or runtime errors; let $N_{\text{total}}$ denote the total number of generated programs and $N_{\text{invalid}}$ the number of failures. The invalid ratio is defined as
\begin{equation}
    \text{Inv.} = \frac{N_{\text{invalid}}}{N_{\text{total}}}.
\end{equation}

A lower invalid ratio indicates more reliable program generation.

For VLM evaluation, we report the average score across the test set. For r-IoU, CD, and r-CD, we report both the mean and median values. The \textbf{mean} captures the overall expected performance and is sensitive to outliers, reflecting the aggregate behavior of the model. The \textbf{median} is more robust to extreme cases and better characterizes the typical per-instance performance.

To avoid overestimating methods with high evaluation failure rates, we adopt \textbf{validity-aware evaluation} in Table~\ref{tab:cadfusion-hard-main}. Metrics are computed over the entire test set rather than conditioning only on successfully evaluated samples, which would otherwise bias results toward methods that fail more frequently but perform well on a small subset of valid outputs. For samples without valid outputs, we apply a conservative imputation strategy using tail quantiles estimated from valid samples: the 90th percentile for error metrics (CD and r-CD, where larger values indicate worse performance) and the 10th percentile for reward-like metrics (r-IoU and VLM score, where smaller values indicate worse performance). This approximates near-worst observed behavior while avoiding instability from extreme outliers, yielding a more faithful estimate of expected model performance.

\subsection{Baseline Reproduction Details}
\label{app:baseline_details}

For Text2CAD\cite{khan2024text2cad}, Cadrille\cite{2025cadrille}, CAD-Coder\cite{guan2025cadcoder}, and CadFusion\cite{wang2025cadfusion}, we strictly use their publicly released code and pretrained model weights. For CADCodeVerify\cite{cadcodeverify}, due to the absence of official open-source code, we re-implemented the method based on the descriptions and prompts provided in the original paper.

For Vanilla, CoT, and our proposed framework across different backbone models, we set the temperature to 0 and max\_tokens to 10000. For Qwen3.5-27B, the “thinking mode” is disabled, while for GPT-5, the reasoning\_effort is set to minimal. All other parameters remain at their default settings.

During VLM evaluation, we use GPT-5 with a temperature of 0, max\_tokens set to 4,096, and reasoning\_effort set to minimal. All remaining parameters are kept at their default values.

All of our experiments are conducted on A100 GPUs, and the closed-source models are accessed through official channels.

\subsection{Additional Settings for Diagnostic Experiments}
\label{app:diagnostic_settings}

This section provides detailed settings for the diagnostic experiments in Section~\ref{sec:ablation} and Section~\ref{sec:analysis}.
These experiments are designed to analyze the contribution of individual components in \textsc{ExpConCAD}, rather than to provide another full benchmark comparison against prior methods.

For the main comparison in Section~\ref{sec:main_results}, we follow the penalized evaluation protocol described in Section~\ref{sec:exp_setup}, where non-renderable outputs are included in the reported metrics.
For the diagnostic experiments, we compute $\mathrm{VLM}$ on successfully rendered outputs.
For the ablation study, we additionally report $\mathrm{IR}$ to show how verification and other components affect renderability.
For the two design analyses in Section~\ref{sec:analysis}, the compared variants have similar renderability, with $\mathrm{IR}$ differences within 3 percentage points.
Therefore, we report $\mathrm{VLM}$ as the primary diagnostic metric to focus on text-shape consistency.

\paragraph{Settings for the Ablation Study}
\label{app:ablation_settings}

This setting corresponds to the ablation study in Section~\ref{sec:ablation}.
We evaluate all variants on \textsc{CadFusion-Hard}, using the same evaluation split as in the main experiments.
We use Qwen3.5-27B and GPT-5 as backbones, following the same generation settings as the main comparison.

The ablation progressively adds the major components of \textsc{ExpConCAD}.
\textit{Prompt} denotes a strong task-specific prompting baseline.
\textit{+Verif.} adds execution-based verification and revision to reduce non-renderable outputs.
\textit{+CSU} further introduces CAD Construction Structure Understanding, which identifies the relevant CAD elements, operations, and constraint scopes.
\textit{+SCC} adds Spatial Constraint Completion over the recovered construction structure.
\textit{Ours} further incorporates experience-enhanced SCC.

For this diagnostic experiment, we report $\mathrm{VLM}$ on successfully rendered outputs and $\mathrm{IR}$ separately.
This allows us to analyze both text-shape consistency and renderability when progressively adding framework components.
The numerical results are reported in Table~\ref{tab:ablation-cadfusion-hard}.

\paragraph{Settings for Analyzing Construction Structure Understanding}
\label{app:csu_scc_settings}

This setting corresponds to the first design analysis in Section~\ref{sec:analysis}, which studies whether Construction Structure Understanding provides a better basis for Spatial Constraint Completion.
We conduct this experiment on \textsc{CadFusion-Hard} with the same evaluation split used in the main experiments.
We evaluate both Qwen3.5-27B and GPT-5.

We compare three settings.
The first is a verification-only baseline, which uses execution-based verification and revision but does not explicitly perform Construction Structure Understanding or Spatial Constraint Completion.
The second performs direct Spatial Constraint Completion from the underspecified shape description, without first recovering the CAD construction structure.
The third follows the proposed design, where the model first performs Construction Structure Understanding and then completes spatial constraints over the recovered constraint scopes.

This experiment isolates whether SCC should be performed directly from the shape-level description or based on the recovered construction structure.
Since the compared variants have similar renderability, with $\mathrm{IR}$ differences within 3 percentage points, we report $\mathrm{VLM}$ as the primary metric.

\paragraph{Settings for Analyzing Experience Memory}
\label{app:exp_memory_settings}

This setting corresponds to the second design analysis in Section~\ref{sec:analysis}, which studies how the size and quality of experience memory affect Spatial Constraint Completion.
We conduct this experiment on \textsc{CadFusion-Hard} with the same evaluation split used in the main experiments.
We evaluate Qwen3.5-27B and GPT-5 as backbones.

For each backbone, we keep the overall \textsc{ExpConCAD} pipeline fixed and vary only the size of the experience memory.
The experience memory is constructed from prior CAD construction cases, and the number of stored experience items follows the settings shown in Figure~\ref{fig:exp_size_analysis}.
For Qwen3.5-27B, the compact memory contains 50 experience items constructed from 45 source cases.
For GPT-5, the largest memory contains 500 experience items constructed from 158 source cases.
During inference, retrieved experiences are used only to guide Spatial Constraint Completion, while the benchmark evaluation split remains unchanged.

This experiment analyzes whether additional experience improves SCC and whether the benefit comes from memory size alone or from the granularity and relevance of the stored experience.
Since $\mathrm{IR}$ remains comparable across memory-size settings, with differences within 3 percentage points, we report $\mathrm{VLM}$ as the primary metric.

\paragraph{Settings for Cross-Dataset Generalization Analysis}
\label{app:generalization_settings}

This setting corresponds to the generalization analysis in Section~\ref{sec:generalization}.
To study cross-dataset transfer, we conduct an additional experiment with Qwen3.5-27B on two hard sets, \textsc{CadFusion-Hard} and \textsc{Text2CAD-Hard}.
\textsc{CadFusion-Hard} is closer to the source distribution used by \textsc{CadFusion} and by our experience construction, while \textsc{Text2CAD-Hard} serves as an unseen cross-dataset test set.

\textsc{CadFusion} learns an end-to-end mapping from \textsc{CadFusion}-side samples.
In contrast, our framework explicitly completes implicit spatial constraints by first understanding the CAD construction structure, and further enhances Spatial Constraint Completion with an experience memory constructed only from \textsc{CadFusion}-side source cases.
Therefore, this experiment evaluates whether end-to-end learned generation patterns, explicit construction-aware constraint completion, and retrieved constraint-completion experiences can transfer to unseen shape-description distributions.
We report $\mathrm{VLM}$ as the main metric in Section~\ref{sec:generalization}, and provide full results on Qwen3.5-27B in Appendix~\ref{app:text2cadhard-results}.

\section{Supplementary Experimental Analysis}\label{app:supp-analysis}

\subsection{Results under the Conventional Valid-Output Protocol}
\label{app:main_res}

\begin{table*}[t]
    \centering
    \footnotesize
    \setlength{\tabcolsep}{2.8pt}
    \renewcommand{\arraystretch}{1.10}
    
    \begin{adjustbox}{width=0.95\textwidth}
    \begin{tabular}{lcccccccc}
        \toprule
        \multirow{2}{*}{\textbf{Method}} & \multicolumn{8}{c}{\textbf{CadFusion-Hard}} \\
        \cmidrule(lr){2-9}
        & \textbf{VLM$\uparrow$}
        & \textbf{m. r-IoU$\uparrow$}
        & \textbf{md. r-IoU$\uparrow$}
        & \textbf{m. CD$\downarrow$}
        & \textbf{md. CD$\downarrow$}
        & \textbf{m. rCD$\downarrow$}
        & \textbf{md. rCD$\downarrow$}
        & \textbf{Inv.$\downarrow$} \\
        \midrule
        \rowcolor{lightgrayrow} \multicolumn{9}{c}{\textit{Trained Text-to-CAD Models}} \\
        Text2CAD & 1.88 & 0.182 & 0.140 & 104.983 & 90.268 & 70.634 & 51.074 & \textbf{0.00} \\
        Cadrille & 1.18 & 0.043 & 0.022 & 151.521 & 138.824 & 79.800 & 70.911 & 5.00 \\
        CAD-Coder & 1.57 & 0.290 & 0.263 & 65.237 & 49.155 & 37.554 & 26.538 & 3.00 \\
        CadFusion & 4.90 & 0.343 & 0.331 & 52.149 & 42.988 & 24.631 & 17.924 & 23.50 \\
        \midrule
        \rowcolor{lightgrayrow} \multicolumn{9}{c}{\textit{Qwen3.5-27B}} \\
        Vanilla & 4.68 & \underline{0.356} & 0.327 & 52.358 & 36.661 & 23.958 & \textbf{13.966} & 50.50 \\
        CoT & 5.18 & 0.353 & 0.283 & 52.213 & \textbf{31.758} & \underline{21.981} & 16.007 & 67.00 \\
        CADCodeVerify & 5.60 & \textbf{0.357} & \textbf{0.343} & \textbf{49.321} & 32.743 & 22.625 & 16.397 & 13.50 \\
        \arrayrulecolor{black!35}\hdashline\arrayrulecolor{black}
        \rowcolor{lightbluehighlight} \textbf{Ours w/o Exp.} & \underline{5.67} & 0.345 & \underline{0.336} & \underline{49.682} & \underline{32.589} & 22.192 & 14.201 & \underline{1.00} \\
        \rowcolor{lightbluerow} \textbf{Ours} & \textbf{6.34} & 0.347 & 0.325 & 50.672 & 34.109 & \textbf{21.388} & \underline{14.193} & 2.00 \\
        \midrule
        \rowcolor{lightgrayrow} \multicolumn{9}{c}{\textit{GPT-5}} \\
        Vanilla & 5.89 & 0.372 & 0.344 & 54.463 & 31.330 & 21.525 & 13.202 & 38.50 \\
        CoT & 5.48 & 0.360 & 0.327 & 54.782 & 38.583 & 21.061 & 14.883 & 32.00 \\
        CADCodeVerify & 6.00 & \textbf{0.378} & \textbf{0.372} & \underline{49.202} & \underline{29.224} & 21.357 & \underline{12.964} & 21.00 \\
        \arrayrulecolor{black!35}\hdashline\arrayrulecolor{black}
        \rowcolor{lightbluehighlight} \textbf{Ours w/o Exp.} & \underline{6.16} & 0.362 & \underline{0.353} & 49.954 & 32.062 & \underline{20.330} & 13.239 & \underline{0.50} \\
        \rowcolor{lightbluerow} \textbf{Ours} & \textbf{6.62} & \underline{0.373} & 0.352 & \textbf{47.157} & \textbf{28.680} & \textbf{19.307} & \textbf{11.032} & \underline{0.50} \\
        \bottomrule
    \end{tabular}
    \end{adjustbox}

\caption{Main results on \textsc{CadFusion-Hard}. 
Best and second-best results are shown in \textbf{bold} and \underline{underlined}.}
    \label{tab:cadfusion-hard-main-app}
\end{table*}

We additionally report the original evaluation results under the conventional Text-to-CAD evaluation protocol in Table~\ref{tab:cadfusion-hard-main-app}.
Following prior work, VLM, r-IoU, CD, and rCD are computed only on successfully rendered outputs, while Invalid Ratio is reported separately.
Therefore, these metrics reflect the quality of valid generations, but do not directly penalize methods for failed rendering cases.
This protocol is useful for comparing the geometric and text-shape quality of renderable outputs, whereas the validity-aware results in the main paper evaluate the full generation pipeline by treating invalid outputs as task failures.

\subsection{Results on the Full \textsc{CadFusion}}\label{app:full-cadfusion-results}
\begin{table*}[t]
    \centering
    \footnotesize
    \setlength{\tabcolsep}{2.8pt}
    \renewcommand{\arraystretch}{1.10}
    
    \begin{adjustbox}{width=0.95\textwidth}
    \begin{tabular}{lcccccccc}
        \toprule
        \multirow{2}{*}{\textbf{Method}} & \multicolumn{8}{c}{\textbf{CadFusion}} \\
        \cmidrule(lr){2-9}
        & \textbf{VLM$\uparrow$}
        & \textbf{m. r-IoU$\uparrow$}
        & \textbf{md. r-IoU$\uparrow$}
        & \textbf{m. CD$\downarrow$}
        & \textbf{md. CD$\downarrow$}
        & \textbf{m. rCD$\downarrow$}
        & \textbf{md. rCD$\downarrow$}
        & \textbf{Inv.$\downarrow$} \\
        \midrule
        \rowcolor{lightgrayrow} \multicolumn{9}{c}{\textit{Trained Text-to-CAD Models}} \\
        Text2CAD & 2.25 & 0.210 & 0.150 & 93.000 & 83.910 & 59.860 & 43.950 & \textbf{0.00} \\
        Cadrille & 1.26 & 0.048 & 0.021 & 160.193 & 143.754 & 84.065 & 75.201 & 3.50 \\
        CAD-Coder & 2.18 & 0.307 & 0.272 & 63.518 & 48.638 & 38.183 & 24.449 & 1.89 \\
        CadFusion & 6.22 & 0.384 & 0.349 & 52.901 & 36.485 & 22.450 & 14.845 & 18.59 \\
        \midrule
        \rowcolor{lightgrayrow} \multicolumn{9}{c}{\textit{Qwen3.5-27B}} \\
        Vanilla & 5.93 & \underline{0.376} & 0.344 & 49.639 & 35.362 & 21.864 & \underline{13.170} & 52.00 \\
        CoT & 7.11 & \textbf{0.390} & 0.350 & 50.606 & 36.800 & 21.160 & 14.027 & 62.39 \\
        CADCodeVerify & 6.69 & 0.375 & 0.349 & 51.004 & 37.668 & 21.879 & 14.224 & 13.55 \\
        \arrayrulecolor{black!35}\hdashline\arrayrulecolor{black}
        \rowcolor{lightbluehighlight} \textbf{Ours w/o Exp.} & \underline{7.56} & \underline{0.376} & \textbf{0.363} & \underline{48.087} & \underline{33.296} & \underline{19.653} & \textbf{12.425} & 1.37 \\
        \rowcolor{lightbluerow} \textbf{Ours} & \textbf{7.74} & 0.370 & \underline{0.355} & \textbf{47.651} & \textbf{32.116} & \textbf{19.450} & 13.228 & \underline{1.16} \\
        \midrule
        \rowcolor{lightgrayrow} \multicolumn{9}{c}{\textit{GPT-5}} \\
        Vanilla & 7.46 & \textbf{0.411} & \underline{0.383} & 49.564 & 33.473 & 19.677 & \underline{11.478} & 33.19 \\
        CoT & 7.59 & \underline{0.403} & \textbf{0.391} & 48.406 & \underline{32.293} & 19.383 & 12.282 & 27.42 \\
        CADCodeVerify & 7.34 & 0.391 & 0.365 & 48.725 & 33.905 & 20.336 & 12.987 & 16.38 \\
        \arrayrulecolor{black!35}\hdashline\arrayrulecolor{black}
        \rowcolor{lightbluehighlight} \textbf{Ours w/o Exp.} & \textbf{7.72} & 0.398 & 0.375 & \textbf{46.325} & \textbf{31.833} & \textbf{18.874} & \textbf{10.891} & \underline{0.11} \\
        \rowcolor{lightbluerow} \textbf{Ours} & \underline{7.69} & 0.395 & 0.367 & \underline{47.636} & 33.760 & \underline{19.239} & 11.735 & 0.32 \\
        \bottomrule
    \end{tabular}
    \end{adjustbox}
\caption{Main results on the full \textsc{CadFusion} benchmark. Best and second-best results are shown in \textbf{bold} and \underline{underlined}, respectively.}
    \label{tab:cadfusion-full-main}
\end{table*}

Table~\ref{tab:cadfusion-full-main} shows that our two variants achieve the best VLM scores on the full \textsc{CadFusion} benchmark.
This confirms the effectiveness of the CG--ICC pipeline on the general test set.
However, compared with \textsc{CadFusion-Hard}, experience brings less stable gains.
The reason is that the full benchmark contains many easier cases whose missing constraints can already be completed by the structured pipeline.
Thus, experience improves Qwen3.5-27B from 7.56 to 7.74, but slightly hurts GPT-5, where the stronger base model may not need additional experience guidance for these simpler cases.

\subsection{Results on the \textsc{Text2CAD-Hard}}\label{app:text2cadhard-results}
\begin{table*}[t]
    \centering
    \footnotesize
    \setlength{\tabcolsep}{2.8pt}
    \renewcommand{\arraystretch}{1.10}
    
    \begin{adjustbox}{width=0.95\textwidth}
    \begin{tabular}{lcccccccc}
        \toprule
        \multirow{2}{*}{\textbf{Method}} & \multicolumn{8}{c}{\textbf{Text2CAD-Hard}} \\
        \cmidrule(lr){2-9}
        & \textbf{VLM$\uparrow$}
        & \textbf{m. r-IoU$\uparrow$}
        & \textbf{md. r-IoU$\uparrow$}
        & \textbf{m. CD$\downarrow$}
        & \textbf{md. CD$\downarrow$}
        & \textbf{m. rCD$\downarrow$}
        & \textbf{md. rCD$\downarrow$}
        & \textbf{Inv.$\downarrow$} \\
        \midrule
        \rowcolor{lightgrayrow} \multicolumn{9}{c}{\textit{Trained Text-to-CAD Models}} \\
        Text2CAD & 2.74 & 0.239 & 0.162 & 80.012 & 69.586 & 48.327 & 28.816 & \textbf{0.00} \\
        Cadrille & 1.38 & 0.038 & 0.013 & 170.415 & 151.596 & 92.485 & 87.336 & 3.00 \\
        CAD-Coder & 2.94 & 0.306 & 0.281 & 66.779 & 49.126 & 44.868 & 27.259 & 3.00 \\
        CadFusion & 5.05 & \textbf{0.354} & \textbf{0.312} & 55.036 & 42.341 & \textbf{26.226} & \textbf{17.069} & 14.00 \\
        \midrule
        \rowcolor{lightgrayrow} \multicolumn{9}{c}{\textit{Qwen3.5-27B}} \\
        Vanilla & 6.23 & 0.313 & 0.261 & 62.356 & 46.800 & 39.100 & 26.871 & 52.00 \\
        CoT & 5.62 & 0.322 & 0.282 & 67.111 & 55.798 & 37.583 & 22.415 & 62.00 \\
        CADCodeVerify & \underline{6.29} & 0.305 & 0.259 & 56.602 & 46.764 & 33.698 & 24.362 & 25.00 \\
        \arrayrulecolor{black!35}\hdashline\arrayrulecolor{black}
        \rowcolor{lightbluehighlight} \textbf{Ours w/o Exp.} & 6.25 & 0.321 & 0.292 & \underline{54.197} & \underline{38.604} & 33.434 & 21.455 & \textbf{0.00} \\
        \rowcolor{lightbluerow} \textbf{Ours} & \textbf{6.54} & \underline{0.331} & \underline{0.306} & \textbf{49.759} & \textbf{35.803} & \underline{30.055} & \underline{19.411} & \underline{0.50} \\
        \bottomrule
    \end{tabular}
    \end{adjustbox}

\caption{
Supplementary results on \textsc{Text2CAD-Hard} for the robustness and experience-transfer analysis in Section~\ref{sec:generalization}.
Best and second-best results are shown in \textbf{bold} and \underline{underlined}.
}
    \label{tab:text2cad-hard-main}
\end{table*}

See in Table~\ref{tab:text2cad-hard-main}.

\subsection{Ablation Study}\label{app:ablation}
\begin{table*}[t]
    \centering
    \footnotesize
    \setlength{\tabcolsep}{2.8pt}
    \renewcommand{\arraystretch}{1.10}
    
    \begin{adjustbox}{width=0.95\textwidth}
    \begin{tabular}{lcccccccc}
        \toprule
        \multirow{2}{*}{\textbf{Method}} & \multicolumn{8}{c}{\textbf{CadFusion-Hard}} \\
        \cmidrule(lr){2-9}
        & \textbf{VLM$\uparrow$}
        & \textbf{m. r-IoU$\uparrow$}
        & \textbf{md. r-IoU$\uparrow$}
        & \textbf{m. CD$\downarrow$}
        & \textbf{md. CD$\downarrow$}
        & \textbf{m. rCD$\downarrow$}
        & \textbf{md. rCD$\downarrow$}
        & \textbf{Inv.$\downarrow$} \\
        \midrule
        \rowcolor{lightgrayrow} \multicolumn{9}{c}{\textit{Qwen3.5-27B}} \\
        Vanilla & 4.68 & \textbf{0.356} & 0.327 & 52.358 & 36.661 & 23.958 & \textbf{13.966} & 50.50 \\
        CAD Prompt & 4.76 & 0.333 & 0.323 & \underline{47.993} & 37.835 & 24.506 & 17.266 & 11.00 \\
        \arrayrulecolor{black!35}\hdashline\arrayrulecolor{black}
        w/o SCC.+CSU & 4.96 & 0.343 & \underline{0.333} & \textbf{46.638} & \underline{33.408} & 22.320 & 15.342 & 3.50 \\
        w/o SCC. & 5.29 & 0.332 & 0.326 & 54.636 & 35.848 & 23.883 & 16.889 & 3.00 \\
        w/o CSU & 5.18 & 0.345 & 0.332 & 51.741 & 34.648 & \textbf{21.015} & 14.320 & 4.00 \\
        \rowcolor{lightbluehighlight} \textbf{w/o Exp.} & \underline{5.67} & 0.345 & \textbf{0.336} & 49.682 & \textbf{32.589} & 22.192 & 14.201 & \textbf{1.00} \\
        \rowcolor{lightbluerow} \textbf{Ours} & \textbf{6.34} & \underline{0.347} & 0.325 & 50.672 & 34.109 & \underline{21.388} & \underline{14.193} & \underline{2.00} \\
        \midrule
        \rowcolor{lightgrayrow} \multicolumn{9}{c}{\textit{GPT-5}} \\
        Vanilla & 5.89 & 0.372 & 0.344 & 54.463 & 31.330 & 21.525 & 13.202 & 38.50 \\
        CAD Prompt & 5.4 & \underline{0.380} & \underline{0.359} & \textbf{45.882} & 38.232 & 20.466 & 12.048 & 8.00 \\
        \arrayrulecolor{black!35}\hdashline\arrayrulecolor{black}
        w/o SCC+CSU & 5.61 & \textbf{0.387} & \textbf{0.374} & 50.075 & 29.105 & 19.683 & 13.822 & 1.50 \\
        w/o SCC. & 5.92 & 0.379 & 0.349 & 49.270 & \textbf{28.043} & \textbf{18.696} & 12.936 & 2.50 \\
        w/o CSU. & 5.56 & 0.379 & 0.355 & 49.344 & 29.563 & 19.045 & \underline{11.424} & \textbf{0.00} \\
        \rowcolor{lightbluehighlight} \textbf{w/o Exp.} & \underline{6.16} & 0.362 & 0.353 & 49.954 & 32.062 & 20.330 & 13.239 & \underline{0.50} \\
        \rowcolor{lightbluerow} \textbf{Ours} & \textbf{6.62} & 0.373 & 0.352 & \underline{47.157} & \underline{28.680} & \underline{19.307} & \textbf{11.032} & \underline{0.50} \\
        \bottomrule
    \end{tabular}
    \end{adjustbox}

\caption{Ablation study on \textsc{CadFusion-Hard}. 
\textbf{Vanilla} uses a generic prompt. 
\textbf{CAD Prompt} denotes our CAD-tailored prompt design without the full structured pipeline. 
\textbf{Ours} is the full framework. 
\textbf{w/o Exp.} removes reusable experience for constraint completion. 
\textbf{w/o SCC} removes spatial constraint completion. 
\textbf{w/o CSU} removes construction structure understanding. 
\textbf{w/o SCC+CSU} removes both spatial constraint completion and construction structure understanding. 
Best and second-best results are shown in \textbf{bold} and \uline{underlined}, respectively.}
    \label{tab:ablation-cadfusion-hard}
\end{table*}

Table~\ref{tab:ablation-cadfusion-hard} shows that \textsc{ExpConCAD} consistently achieves the best VLM score on both Qwen3.5-27B and GPT-5.
Compared with Vanilla and CAD Prompt, the full framework brings clear improvements, indicating that a structured generation pipeline is more effective than prompt design alone.
Removing CG or ICC weakens performance, showing that both CAD-context grounding and implicit constraint completion are useful for vague-shape-level Text-to-CAD.
Moreover, the gap between \textbf{w/o Exp.} and \textbf{Ours} shows that reusable experience further improves the quality of constraint completion.

\subsection{Additional Analysis of Experience in \textsc{ExpConCAD}}
\label{app:expconcad_cases}

This section provides additional analysis of the experience mechanism in \textsc{ExpConCAD}.
We show how different backbones produce experiences at different granularities, and how this affects the usability and retrieval effectiveness of the experiences.

\begin{figure*}[p]
    \centering
    \includegraphics[width=\textwidth]{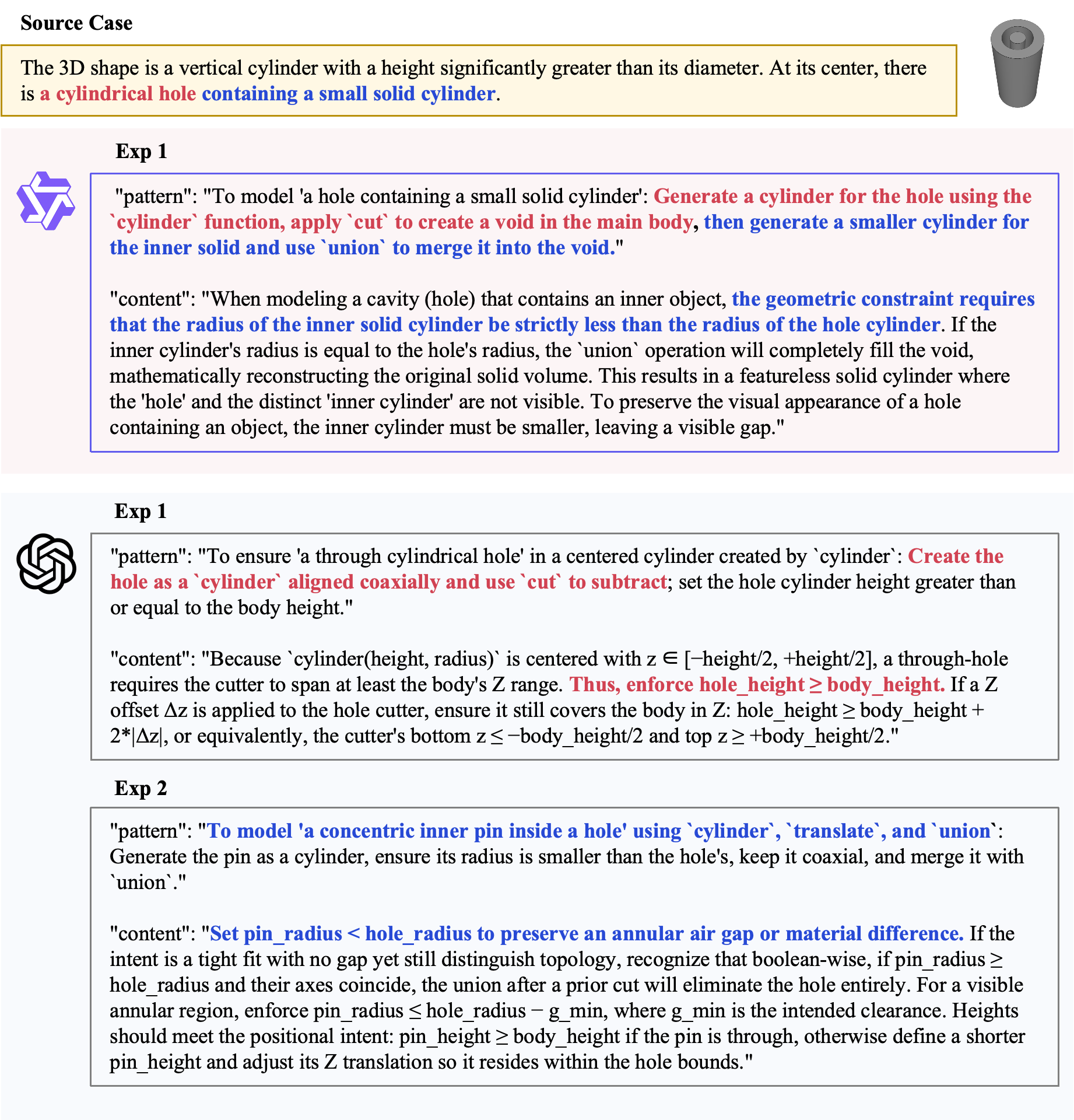}
\caption{
Comparison of experiences generated by Qwen3.5-27B and GPT-5.
}
    \label{fig:exp_compare}
\end{figure*}

Figure~\ref{fig:exp_compare} illustrates the differences in experiences summarized by Qwen3.5-27B and GPT-5 for the same source case. The source case is described as: "The 3D shape is a vertical cylinder with a height significantly greater than its diameter. At its center, there is a cylindrical hole containing a small solid cylinder." It involves two distinct patterns: (1) constructing a cylindrical hole; (2) modeling a small cylinder inside the hole.

Qwen3.5-27B summarizes a single experience whose pattern simultaneously covers both constructing the hole and inserting the cylinder. However, its content only specifies the constraint for inserting the cylinder, namely that the cylinder's radius should be smaller than the hole's radius. Consequently, even when this experience is retrieved, it cannot provide guidance for constructing the hole, limiting its usability for tasks that require hole creation.

In contrast, GPT-5 generates two separate experiences, each aligned with a specific pattern. The first experience focuses on constructing the hole, specifying in its content that the hole height must exceed the body height. The second experience focuses on inserting the cylinder, explicitly detailing the corresponding constraints. This separation allows GPT-5 to provide targeted guidance for each pattern. 

As a result, GPT-5 offers more generalizable and reusable experiences across different CAD contexts. Its fine-grained, pattern-specific experiences improve retrieval accuracy and task performance, demonstrating greater versatility compared to the coarse, case-level guidance produced by Qwen3.5-27B.

\subsection{Token Consumption Analysis}
\label{app:tokens}
\begin{figure*}[h]
    \centering
    \includegraphics[width=\textwidth]{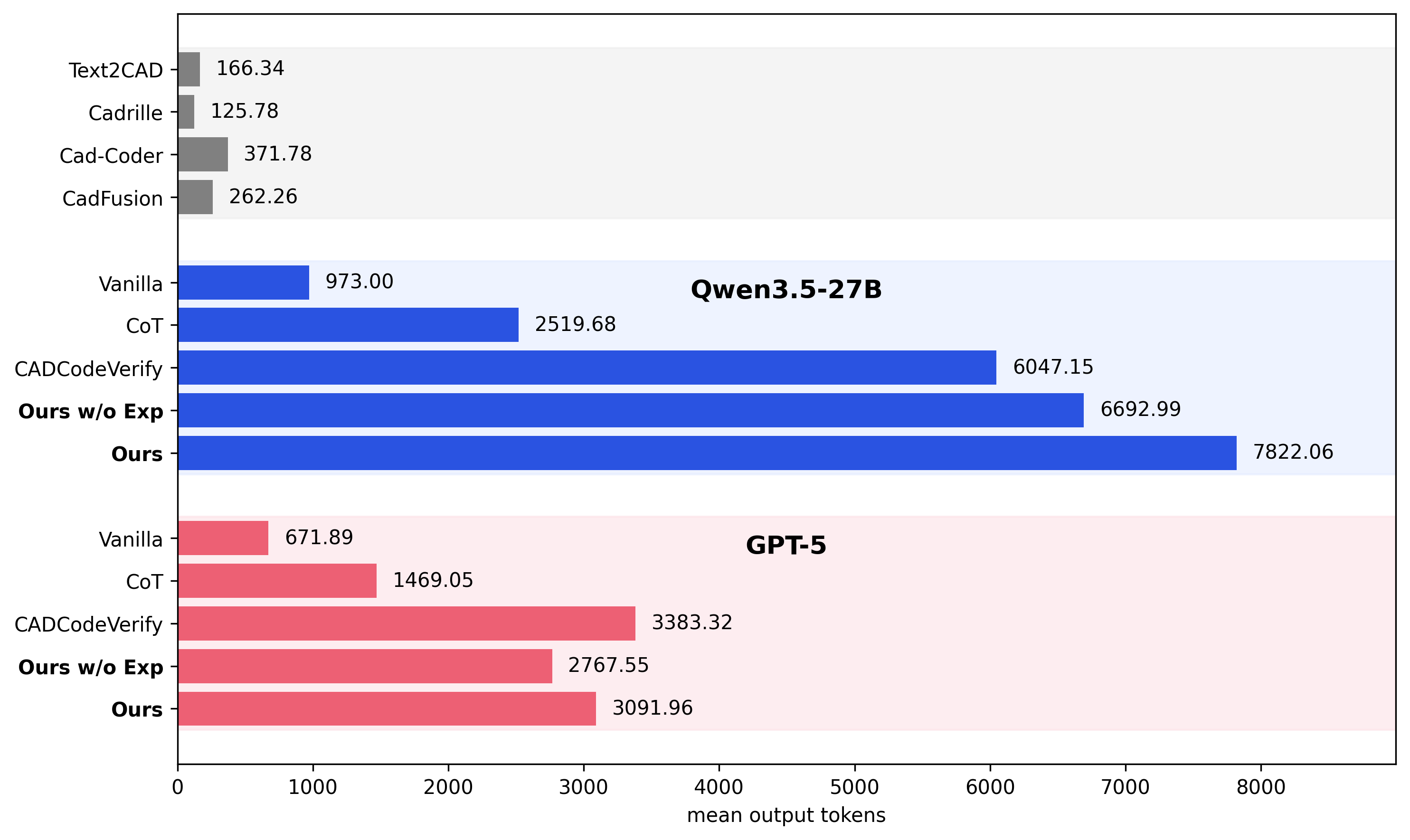}
\caption{
Comparison of mean output tokens.
}
    \label{fig:toke_comparison}
\end{figure*}

We report the token consumption of different methods in Figure~\ref{fig:toke_comparison}. Although our method achieves strong performance, this improvement comes at the cost of increased token usage, and the overhead varies across backbone models. 

When using Qwen3.5-27B as the backbone, we observe that the model occasionally falls into repetitive generation loops, repeatedly producing duplicated content until the context window is exhausted. This behavior appears to be associated with limitations in the model's reasoning capability and is more likely to occur on challenging cases requiring complex reasoning. As a result, the average token consumption is significantly inflated. Specifically, under the Qwen3.5-27B backbone, repetitive generation occurs in 8.5\% of the samples for \textit{Ours w/o Exp} and 10\% for \textit{Ours}. 

In contrast, such context-saturating repetition is absent when GPT-5 is adopted as the backbone. Moreover, under the GPT-5 backbone, our framework not only outperforms CADCodeVerify in task performance but also achieves lower token consumption.
\label{sec:appendix}

\end{document}